\documentclass[acmtog, balance = false]{acmart}
\acmSubmissionID{344}

\acmJournal{TOG}
\setcitestyle{square}
\usepackage{lipsum}
\usepackage{url}

\usepackage{wrapfig}
\usepackage{nicefrac}
\usepackage{mathtools}
\usepackage{graphicx}
\usepackage{booktabs} 
\usepackage{combelow} 
\usepackage[ruled,vlined,linesnumbered]{algorithm2e}
\usepackage{tabularx} 
\usepackage{colortbl} 

\usepackage{manfnt} 

\usepackage{amsmath}
\usepackage{amsfonts}
\usepackage{amsbsy}

\usepackage{listings} 
\usepackage{courier}
\definecolor{mygreen}{rgb}{0,0.6,0}
\definecolor{mygray}{rgb}{0.5,0.5,0.5}
\definecolor{mymauve}{rgb}{0.58,0,0.82}
\definecolor{superlightgray}{RGB}{240,240,240}
\usepackage[mathletters]{ucs}
\usepackage[utf8x]{inputenc}

\definecolor{derekBlue}{RGB}{144,210,236}
\definecolor{derekTableBlue}{RGB}{189,235,252}
\definecolor{iglGreen}{RGB}{153,203,67}
\definecolor{coralRed}{RGB}{250,114,104}
\definecolor{gray}{RGB}{180,180,180}
\definecolor{orange}{RGB}{255,165,0}
\definecolor{TechnionBlue}{RGB}{8,33,78}
\definecolor{lightgray}{gray}{0.65}

\SetCommentSty{commentFont}
\SetNlSty{textbf}{}{.}

\newcommand{\refequ}[1] {Eq.~\eqref{equ:#1}}

\newcommand{\reffig}[1] {Fig.~\ref{fig:#1}}
\newcommand{\reffignum}[1] {\ref{fig:#1}}
\newcommand{\reftab}[1] {Table~\ref{tab:#1}}
\newcommand{\refsec}[1] {Sec.~\ref{sec:#1}}
\newcommand{\refapp}[1] {App.~\ref{app:#1}}

\newcommand{\R}{\mathbb{R}}
\newcommand{\lipc}{c}

\newcommand{\softplus}{\textit{softplus}\,}

\newcommand{\vecFont}[1]{\mathsf{#1}}

\def\vb{{\vecFont{b}}}

\def\vp{{\vecFont{p}}}

\def\vt{{\vecFont{t}}}

\def\vx{{\vecFont{x}}}
\def\vy{{\vecFont{y}}}
\def\vz{{\vecFont{z}}}

\newcommand{\matFont}[1]{\mathsf{#1}}

\def\mI{{\matFont{I}}}
\def\mJ{{\matFont{J}}}

\def\mM{{\matFont{M}}}

\def\mV{{\matFont{V}}}
\def\mW{{\matFont{W}}}

\newcommand{\update}[1]{#1}
\newcommand{\updateNew}[1]{#1}

\AtBeginDocument{%
  \providecommand\BibTeX{{%
    \normalfont B\kern-0.5em{\scshape i\kern-0.25em b}\kern-0.8em\TeX}}}

\copyrightyear{2022} 
\acmYear{2022} 
\setcopyright{acmcopyright}\acmConference[SIGGRAPH '22 Conference Proceedings]{Special Interest Group on Computer Graphics and Interactive Techniques Conference Proceedings}{August 7--11, 2022}{Vancouver, BC, Canada}
\acmBooktitle{Special Interest Group on Computer Graphics and Interactive Techniques Conference Proceedings (SIGGRAPH '22 Conference Proceedings), August 7--11, 2022, Vancouver, BC, Canada}
\acmPrice{15.00}
\acmDOI{10.1145/3528233.3530713}
\acmISBN{978-1-4503-9337-9/22/08}

\begin{document} 

\title{Learning Smooth Neural Functions via Lipschitz Regularization}

\author{Hsueh-Ti Derek Liu}
\affiliation{%
  \institution{University of Toronto}
  \country{Canada}
  }
\email{hsuehtil@cs.toronto.edu}

\author{Francis Williams}
\affiliation{%
  \institution{NVIDIA}
  \country{USA}
  }
\email{francis@fwilliams.info}

\author{Alec Jacobson}
\affiliation{%
  \institution{University of Toronto \& Adobe Research}
  \country{Canada}
  }
\email{jacobson@cs.toronto.edu}

\author{Sanja Fidler}
\affiliation{%
  \institution{University of Toronto \& NVIDIA}
  \country{Canada}
  }
\email{fidler@cs.toronto.edu}

\author{Or Litany}
\affiliation{%
  \institution{NVIDIA}
  \country{USA}
  }
\email{or.litany@gmail.com} 
\renewcommand{\shortauthors}{Liu, et al.}






\begin{teaserfigure}
  \includegraphics[width=\linewidth]{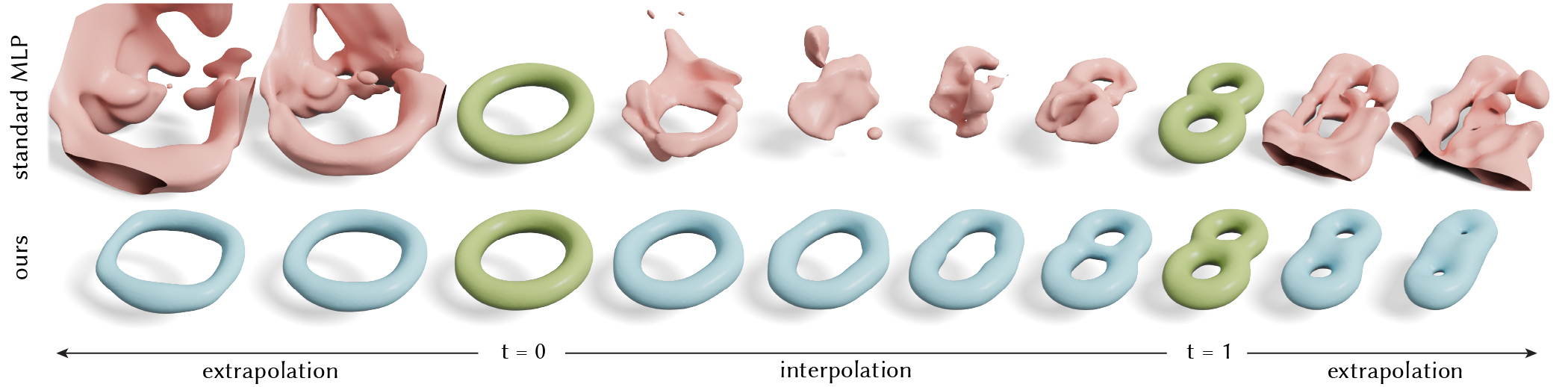}
  \vspace{-10pt}
  \caption{We \update{fit} neural networks to the signed distance field of a torus when the latent code $t = 0$ and a double torus when $t = 1$ (green). Our Lipschitz multilayer perceptron (MLP) achieves smooth interpolation and extrapolation results (blue) when changing $t$, while the standard MLP fails (red).}
  \label{fig:teaser} 
  \vspace{5pt}
\end{teaserfigure} 

\begin{abstract}
Neural implicit fields have recently emerged as a useful representation for 3D shapes. These fields are commonly represented as neural networks which map latent descriptors and 3D coordinates to implicit function values. 
The latent descriptor of a neural field acts as a deformation handle for the 3D shape it represents. Thus, smoothness with respect to this descriptor is paramount for performing shape-editing operations. In this work, we introduce a novel regularization designed to encourage smooth latent spaces in neural fields by penalizing the upper bound on the field's Lipschitz constant.
Compared with prior Lipschitz regularized networks, ours is computationally fast, can be implemented in four lines of code, and requires minimal hyperparameter tuning for geometric applications. We demonstrate the effectiveness of our approach on shape interpolation and extrapolation as well as partial shape reconstruction from 3D point clouds, showing both qualitative and quantitative improvements over existing state-of-the-art and non-regularized baselines. 
%
\end{abstract}

\maketitle
\section{Introduction}\label{sec:intro}

Neural Fields have become a popular representation for shapes in geometric learning tasks. A neural field is an implicit function encoded as a neural network which maps input 3D coordinates to scalar values (for example signed distances).
In many tasks, these networks are conditioned on an additional \emph{shape latent code} which is learned from a large corpus of shapes and acts as knob to deform the shape encoded by the neural field. 
Thus, smoothness with respect to the latent descriptor of a neural field is a desirable property to encourage well behaved deformations.
%

\begin{figure}
  \begin{center}
  \includegraphics[width=1\linewidth]{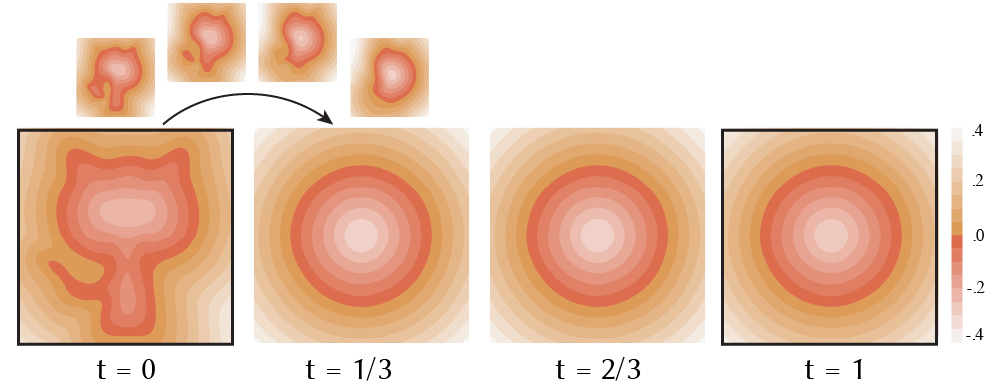}
  \end{center}
  \vspace{-5pt}
  \caption{We \update{fit} a multilayer perceptron to fit the signed distance functions (SDFs) of a cat shape and a circle at $t = 0$ and $t = 1$ respectively. In addition, we minimize the Dirichlet energy of $t$ at $t = \nicefrac{1}{3}, \nicefrac{2}{3}$. While the network finds a smooth solution at those sample time steps, it still has non-uniform change beyond the samples, such as between $0 ≤ t ≤ \nicefrac{1}{3}$.}
  \label{fig:dirichlet}
\end{figure}

There are many traditional ways to encourage a function to posses some notion of smoothness. However we find that such classical approaches are not applicable to obtaining a smooth latent space of neural fields. For example in \reffig{dirichlet}, we minimize the Dirichlet energy defined over the latent space, but the neural network still possesses non-smooth behavior outside the training set.

In this work, we focus on encouraging smoothness with respect to the latent parameter of a neural field. Since neural fields are continuous by construction, we use the \emph{Lipschitz bound} as a metric for smoothness of the latent space. This notion of smoothness is define over the entire space. Thus, it encourages smoothness even away from the training set (\reffig{teaser}).
While Lipschitz constrained networks have been proposed before (see \refsec{relatedwork}), they are not readily applicable to geometric applications.  
%
In particular, they require pre-determining the Lipschitz bound, which is unknown in advanced and highly input dependent (see \reffig{manual_lipc_large}). Therefore, to use prior Lipschitz architectures one has to perform extensive per-shape hyperparameter tuning to find a reasonable Lipschitz constant.  

We therefore propose a novel \emph{smoothness} regularizer to minimize a learned Lipschitz bound on the latent vector of a neural field. 
Our method is extremely simple and effective: one only needs to add a weight normalization layer and augment the loss function with a simple regularization term encouraging small Lipschitz. Unlike previous approaches, our method can perform high quality deformations on latent spaces learned with as few as two shapes. We demonstrate the effectiveness of our method on the tasks of shape interpolation and extrapolation (\refsec{interpolation_extrapolation}), robustness to adversarial inputs (\refsec{robustness}), and shape completion from partial point-clouds (\refsec{testtime}), in which we oupterform past methods both qualitatively and quantitatively.

\section{Related Work}\label{sec:relatedwork}
We focus our discussion on how learning-based methods encourage smoothness and methods similar in methodology. For an overview on neural fields, please refer to \cite{neuralfieldsurvey}.

\begin{figure}
  \begin{center}
  \includegraphics[width=1\linewidth]{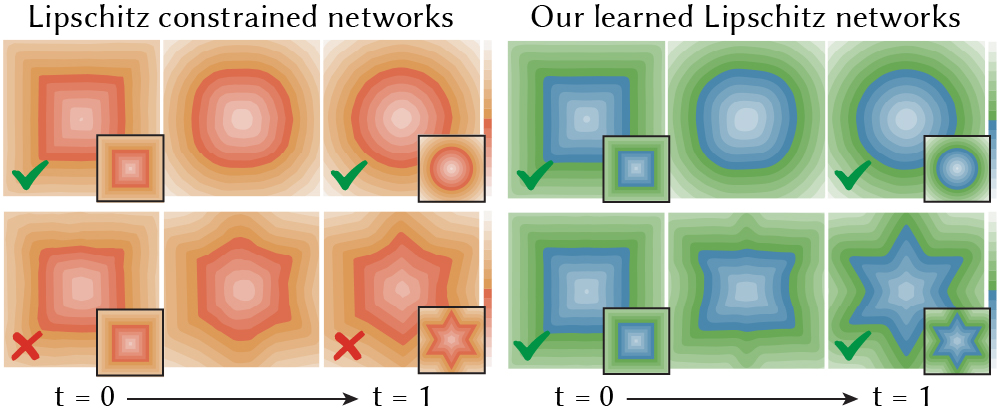}
  \end{center}
  \vspace{-5pt}
  \caption{\update{Different tasks require different Lipschitz constants. We manually specify the same Lipschitz constant on the top two interpolation examples. One of them is sufficient \updateNew{(top left)} to fit the ground truth (black), but the other one is not \updateNew{(bottom left)}. In contrast, our method learns a suitable Lipschitz constant for each task \updateNew{(right)} with the same hyperparameter.}}
  \label{fig:manual_lipc_large}
\end{figure}

\paragraph{Geometric Regularizations} 
Many existing approaches rely on classic measures to encourage smoothness in neural 3D mesh processing.
Several methods (e.g., \cite{liu2019soft, WangZLFLJ18, HertzHGC20}) use Laplacian regularization which penalizes the difference between a vertex and the center of mass of its 1-ring neighbors. \citet{KatoUH18} encourage smoothness by encouraging flat dihedral angle between adjacent faces. \citet{WangZLFLJ18, HertzHGC20} penalize edge-lengths and their variance. \citet{RakotosaonaO20} define an isometry regularization in character deformation to obtain area preserving interpolation. Many techniques (e.g., \cite{ChenLGSLJF19}) even use a mixture of these regularizations. 
However, these regularizations often require the input being a manifold triangle mesh. In other representations such as neural fields, regularizations of the level set surface are difficult to be defined.
Previous works also introduced other geometric regularizations for encouraging different properties, such as \cite{GroppYHAL20, WilliamsTBZ21}. But we exclude our discussion on those techniques because they are not directly related to the smoothness of a network function. 

\paragraph{Network Regularizations} 
Given input samples, one can differentiate through a network to obtain derivative information of the network output with respect to these inputs. Then we can encourage smoothness at these input samples by penalizing the norm of the Jacobian \cite{drucker1991double, judy2019jacobian, JakubovitzG18, varga2017gradient, GulrajaniAADC17} or the Hessian \cite{MoosaviDezfooli19}.
If differentiating through the network is undesirable, \citet{Elsner2021Lipschitz} propose to penalize the difference in the output based on the input similarity.
These techniques are effective in obtaining smooth solutions at training samples, but they have no guarantee to obtain a smooth function beyond them. On the contrary, it may even promote non-smooth behavior by squeezing function changes to locations without training samples (see \reffig{dirichlet}). 

In lieu of this, one should use regularization techniques that do not depend on the input to a network, such as penalizing L2 norm \cite{tihonov1963solution} or L1 norm \cite{tibshirani1996regression} of the weight matrices.
Other training techniques can also be used to regularize the network, such as early-stopping \cite{ulyanov2018deep, WilliamsSSZBP19}, dropout \cite{SrivastavaHKSS14}, and learning rate decay \cite{LiWM19a}. 
Applying these techniques can alleviate overfitting, but how they relate to the smoothness of a network function remains an open problem. In our experiments, they produce less smooth results compared to our method (see \reftab{smoothness_quantitative}). 

\begin{figure}
  \begin{center}
  \includegraphics[width=1\linewidth]{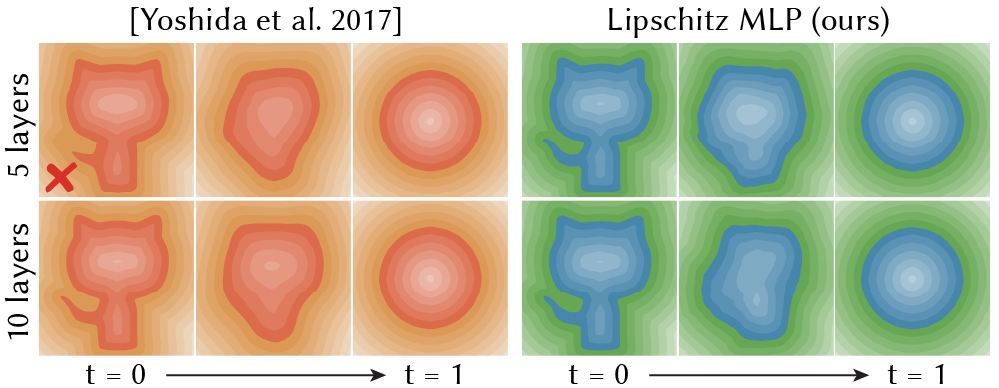}
  \end{center}
  \vspace{-5pt}
  \caption{The spectral norm regularization proposed by \citet{YoshidaM17} \update{is more sensitive to the number of layers}. Therefore, using the same $\alpha$ on a 5-layer and a 10-layer MLP leads to different effects (red). In contrast, our regularization (blue) leads to more consistent results. }
  \label{fig:yoshida_compare}
\end{figure}
\paragraph{Lipschitz Regularizations}
The Lipschitz constant of neural networks has attracted huge attention because of its applications in robustness against adversarial attacks \cite{Adam2018Lipschitz, LiHALGJ19}, better generalization \cite{YoshidaM17}, and Wasserstein generative adversarial networks \cite{arjovsky2017wasserstein}.
Several techniques have been proposed to precisely constraint the Lipschitz constant of a network. \citet{MiyatoKKY18} normalize the weight matrices by dividing each weight matrix by its largest eigenvalue. This \emph{spectral normalization} enforces a neural network to be 1-Lipschitz, a neural network with Lipschitz bound $1$, under the L2 norm. \citet{GoukFPC21} rely on different weight normalization methods to constrain the Lipschitz bound under L1 and L-infinity norms. \citet{CisseBGDU17, AnilLG19} obtain 1-Lipschitz networks by orthonormalizing each weight matrix.
Strictly constraining the Lipschitz constant of a network is not always desirable because it may lead to undesired behavior in the optimization \cite{GulrajaniAADC17, Mihaela2020smoothness}. In response, \citet{Terjek20} propose a regularization to softly encourage a network to be $c$-Lipschitz.
However, these Lipschitz constrained networks often fail to achieve the prescribed Lipschitz bound because the estimated Lipschitz bound is not tight due to the ignorance of activation functions. Several papers complement this subject by proposing more accurate methods to estimate the true Lipschitz constant, such as \cite{VirmauxS18, WengZCSHDBD18, JordanD20}. \citet{AnilLG19} propose a new activation function based on sorting to tighten the estimated Lipschitz bound. 
Unfortunately, the above-mentioned Lipschitz constrained networks require to know the target Lipschitz constant beforehand. This makes them difficult to be deployed to geometry applications because a good Lipschitz constant is unknown, thus leading to extensive hyperparameter tuning (see \reffig{manual_lipc_large}).

%
This inspires some Lipschitz-like regularizations, such as the spectral norm regularization which penalizes the largest eigenvalue of each weight matrix \cite{YoshidaM17}. They also show that adding regularization improves the generalizability and adversarial robustness. But this regularization does not incorporate the fact that the Lipschitz constant grows exponentially with respect to the depth of the network. In practice, it causes difficulties in hyperparameter tuning because changing the number of layers requires to also change the weight on the regularization (see \reffig{yoshida_compare}).

\section{Background in Lipschitz Networks}\label{sec:background}
A neural network $f_\theta$ with parameter $\theta$ is called \emph{Lipschitz continuous} if there exist a constant $\lipc ≥ 0$ such that
\begin{align}
  \underbrace{\| f_\theta(\vt_0) - f_\theta(\vt_1) \|_p}_{\mathclap{\text{change in the output}}} ≤ \lipc\ \underbrace{\| \vt_0 - \vt_1\|_p}_{\mathclap{\text{change in the input}}}
\end{align}
for all possible inputs $\vt_0, \vt_1$ under a $p$-norm of choice. The parameter $\lipc$ is called the \emph{Lipschitz constant}. Intuitively, this constant $\lipc$ bounds how fast this function $f_\theta$ can change. 

As pointed out by several previous papers mentioned in \refsec{relatedwork}, the Lipschitz bound $\lipc$ of an fully-connected network with 1-Lipschitz activation functions (e.g., ReLU) can be estimated via
\begin{align}\label{equ:lipschitz_bound_MLP}
  \lipc = ∏_{i = 1}^L \| \mW_i \|_p,
\end{align}
where $\mW_i$ is the weight matrix at layer $i$ and $L$ denotes the number of layers. This estimate is a loose upper bound due to the ignorance of activation functions, but in practice, optimizing this upper bound is still effective (i.e. \cite{MiyatoKKY18}). 

In the past years, different ways of controlling the Lipschitz bound of a network have been studied. A dominant strategy is to perform weight normalization. For instance, if one wants to enforce the network to be 1-Lipschitz $\lipc = 1$, then one can achieve this by normalizing the weight such that $\| \mW_i \|_p = 1$ after each gradient step during training. 
The normalization scheme depends on the choice of different matrix $p$-norms: 
\begin{gather}\label{equ:different_norms}
  \| \mM \|_2 = \sigma_\text{max}(\mM),\\
  \| \mM \|_1 = \max_j ∑_i |m_{ij}|, \quad
  \| \mM \|_∞ = \max_i ∑_j |m_{ij}|,
\end{gather}
where $\sigma_\text{max}(\mM)$ denotes the maximum eigenvalue of $\mM$.
Thus, when $p=2$, weight normalization consists of rescaling the weight matrix based on its maximum eigenvalue. Popular techniques include spectral normalization based on the power iteration \cite{MiyatoKKY18} and the Bj\"{o}rck Orthonormalization \cite{bjorck1971iterative, AnilLG19}. 
When $p=\infty$ ($p=1$), weights normalization is simply scaling individual rows (columns) to have a maximum absolute row (column) sum smaller than a prescribed bound. 

These matrix norms are also related to each other. Let $\mM$ be a matrix with size $m$-by-$n$, its $2$-norm is bounded by its $1$-norm and $\infty$-norm in the following relationships
\begin{align}
  & \frac{1}{\sqrt{n}} \| \mM \|_\infty  ≤ \|\mM \|_2 ≤ \sqrt{m} \| \mM \|_\infty \\
  & \frac{1}{\sqrt{m}} \| \mM \|_1 ≤ \|\mM \|_2 ≤ \sqrt{n} \| \mM \|_1.
\end{align}
This implies that optimizing the Lipschitz bound under a particular choice of norms will effectively optimize the bound measured by the other norms.
%
%
One could also consider the entry-wise matrix norm $\| \mM \|_{p,q}$ (see \cite{horn2012matrix}). But we leave the exploration of the most effective strategy as future work. 


\begin{figure}
  \begin{center}
  \includegraphics[width=1\linewidth]{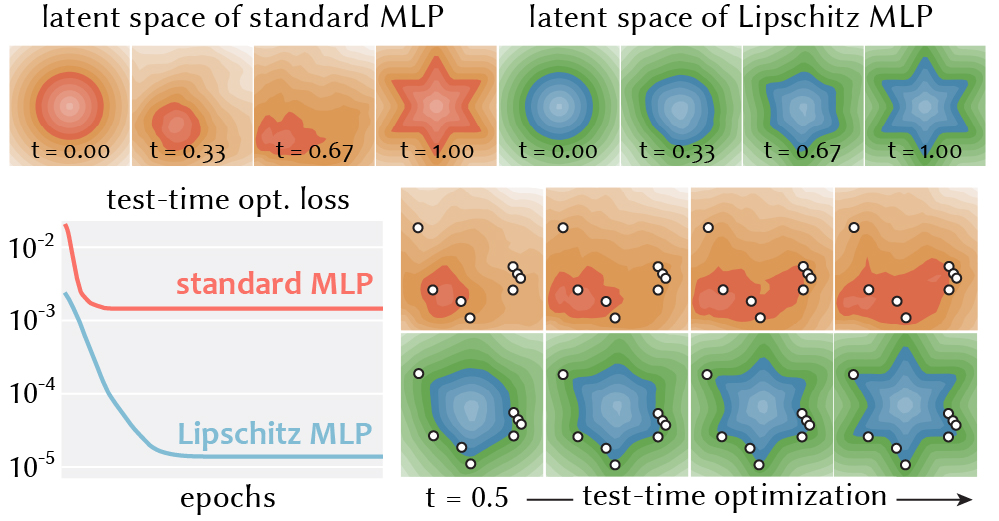}
  \end{center}
  \vspace{-5pt}
  \caption{We demonstrate the importance of smooth latent space on a toy test-time reconstruction of a 2D point cloud. We train a standard MLP and our Lipschitz MLP to interpolate a circle ($t=0$) to a star ($t=1$). Then we sample points on the zero-isoline of the star (white points) and optimize the latent code to fit those points (with initial $\vt = 0.5$). Our smooth latent space enables latent optimization to find the correct solution (bottom blue) while the non-smooth one suffers from poor local minimum (bottom red).}
  \label{fig:toy_testtime_adam}
  \vspace{-10pt}
\end{figure}
\section{Method}\label{sec:method}
Throughout we use $f_\theta(\vx, \vt)$ to denote the forward model of an implicit shape parameterized by a neural network, namely a mapping from a tuple of a location $\vx ∈ \R^d$ in $d$-dimensional space and a latent code $\vt ∈ \R^{|\vt|}$ to $\R$. $f_\theta$ has parameters $θ = \{ \mW_i, \vb_i\}$ containing weights $\mW_i$ and biases $\vb_i$ of each layer $i$. 

Our goal is to train a neural network that is smooth with respect to its latent code $\vt$. This property is important for shape editing and in applications requiring a well-structured latent space in which a small change in $\vt$ results in a small change to the output (see \reffig{toy_testtime_adam}). 

A straightforward idea is to augment the loss function with some smoothness regularizations, such as the Dirichlet energy. Specifically, one would draw a bunch of samples points $\vt_j$ in the latent space and turn the original loss function $\mathcal{L}$ into
\begin{align}\label{equ:dirichlet_regularization}
  \mathcal{J}(\theta) = \mathcal{L}(\theta) + \alpha\ \sum_j \| \frac{∂ f_\theta}{∂ \vt}(\vx, \vt_j)\|^2.
\end{align}
Although being effective in encouraging a smooth neural field $f_\theta$ with respect to the change in latent code at the sampled locations $\vt_j$, it often results in non-smooth behavior elsewhere. For instance, in \reffig{dirichlet}, we apply the Dirichlet regularization to a toy task: interpolating between two neural SDFs conditioned on two latent codes, $\vt=0$ and $\vt=1$, with 1D latent dimension. 
In this example, we minimize the Dirichlet energy at $\vt = \nicefrac{1}{3}, \nicefrac{2}{3}$. The network is able to find a perfectly smooth (constant) solution at the \update{sampled} $\vt$s, but it squeezes all the changes at the very beginning and results in non-smooth behavior $0 < \vt < \nicefrac{1}{3}$. This issue is even more troublesome when the latent dimension is large and \update{sampling densely} is intractable.
A more desirable approach is to guarantee smoothness for \emph{all} possible latent inputs without the need to densely sample the latent space. 

Our main idea is to define the smoothness energy solely based on network parameters (i.e. weights of a neural network) regardless of the inputs. 
One promising solution is to encourage \emph{Lipschitz continuity} with respect to the inputs, in our case the latent code $\vt$, and use its Lipschitz constant $\lipc$ as a proxy for smoothness. Specifically, we want the network to satisfy
\begin{align}\label{equ:lipschitz_def}
  \| f_\theta(\vx, \vt_0) - f_\theta(\vx, \vt_1) \|_p ≤ \lipc\ \| \vt_0 - \vt_1\|_p
\end{align}
for all possible combinations of $\vx, \vt_0, \vt_1$. As the upper bound of the Lipschitz constant $\lipc = ∏_i \| \mW_i \|_p$ only depends on the weight matrices $\mW_i$, $\lipc$ is independent to the choice of inputs. Therefore, by decreasing $\lipc$, one guarantees smoothness everywhere even beyond the training set \update{(see \reffig{teaser})}. 
%

To decrease the Lipschitz constant and encourage smoothness, we present a new regularization. The key idea is to treat the Lipschitz constant of a network as a learnable parameter and minimize it, instead of a pre-determined value (e.g., \cite{MiyatoKKY18}).
There are many possible ways one can formulate such a regularization and there is not a single formulation that is uniformly the best. We first present our recommended solution and defer the comparison with alternative formulations in \refsec{comparisons}.

\subsection{Lipschitz Multilayer Perceptron}
%
%
The first question is to choose a $p$-norm to measure the Lipschitz constant \refequ{lipschitz_def}. In our case, we have no restriction on the choice of $p$-norm. We solely want to have a small Lipschitz constant to encourage smooth behavior with respect to the change of the latent code $\vt$. Thus, we simply choose the matrix $\infty$-norm due to its efficiency (see the inset). But if applications require other choices, our approach is also applicable.

\begin{wrapfigure}[9]{r}{0.4\linewidth}
  \centering
  \vspace*{-1\intextsep}
  \hspace*{-.75\columnsep}
  \includegraphics[width=1.15\linewidth, trim={0mm 2mm 0mm 0mm}]{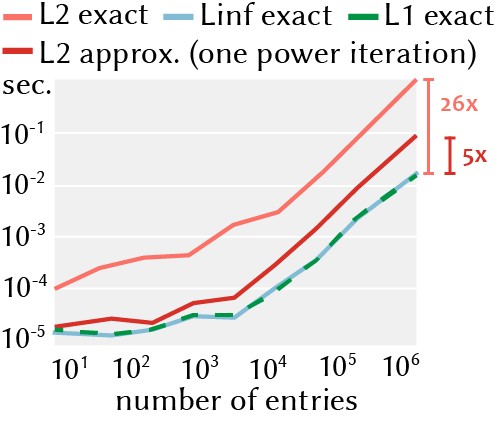}
\end{wrapfigure}
After determining the matrix norm, our method only requires two simple modifications to a standard fully-connected network: adding a weight normalization to each fully connected layer, parameterized by a learnable Lipschitz variable, and a regularization term to encourage an overall small Lipschitz constant that is minimized together with the task loss function. 

\subsubsection{Weight Normalization Layer}\label{sec:normalization_layer}
Our weight normalization shares the same spirit as the other weight normalization methods for accelerating training \cite{SalimansK16} and generalization ability \cite{HuangLLYWL18}. But the key difference is that our normalization is based on the Lipschitz constant of a layer, which is more suitable for obtaining a smooth network.

We augment each layer of an MLP $\vy = σ( \mW_i \vx + \vb_i)$ with a Lipschitz weight normalization layer given a trainable Lipschitz bound $c_i$ for layer $i$ 
\begin{align}\label{equ:normalization_layer}
  \update{\vy = σ( \widehat\mW_i \vx + \vb_i), \quad \widehat\mW_i = \textit{normalization}\,(\mW_i, \softplus(c_i)),}
\end{align}
\update{where the $\softplus(c_i) = \ln(1+e^{c_i})$ is a reparameterization designed to avoid infeasible negative Lipschitz bounds. In most of our cases $c_i ≈ \softplus(c_i)$ because $c_i$ is often a large positive number.}
Due to our choice of using $\infty$-norm, this normalization is efficient and simple: we scale each row of $\mW_i$ to have the absolute value row-sum less than or equal to \update{$\softplus(c_i)$}. If one of the rows already has the absolute value row-sum smaller than $\softplus(c_i)$, then no scaling is performed. 
With this normalization layer, even if the raw weight matrix $\mW_i$ has a Lipschitz constant greater than \update{$\softplus(c_i)$}, this normalization can still guarantee the Lipschitz constant is bounded by \update{$\softplus(c_i)$}. Therefore, we never clip the weights during training. 

\paragraph{Implementation}
\update{Our method can be implemented in a few lines of code.}
Given the weight matrix \texttt{\small Wi} and the per-layer Lipschitz upper bound \texttt{\small ci}, the normalization layer can be implemented in \update{JAX} \cite{jax2018github} as 
\begin{lstlisting}[language=python]
 import jax.numpy as jnp
 def normalization(Wi, softplus_ci): # L-inf norm
   absrowsum = jnp.sum(jnp.abs(Wi), axis=1)
   scale = jnp.minimum(1.0, softplus_ci/absrowsum)
   return Wi * scale[:,None]
\end{lstlisting} 
and each layer of the Lipscthiz MLP is simply 
\begin{lstlisting}[language=python]
 y = sigma(normalization(Wi, softplus(ci))*x + bi)
\end{lstlisting} 
where \texttt{\small\color{blue} sigma} denotes the activation function \update{and \texttt{\small\color{blue} softplus} is the built-in \emph{softplus} function in JAX.}

\update{Although being efficient, using our Lipschitz weight normalization will still increase the training time. For example, in the 2D interpolation task (such as \reffig{manual_lipc_large}), adding our normalization slows down the training from 265.83 epochs per second down to 229.95 epochs per second.}

\update{However, incorporating our regularization will not influence the performance during test time because one can explicitly construct the normalized weight matrix $\widehat{\mW}_i$ by clipping the weight matrix $\mW_i$ with the learned constant $c_i$. Then, one can use the vanilla MLP with the normalized weights $\widehat{\mW}_i$ as their final model.}

\subsubsection{Our Lipschitz Regularization}
The second ingredient is to augment the original loss function $\mathcal{L}$ with a \emph{Lipschitz regularization}. Our Lipschitz regularization is defined simply as the Lipschitz bound of the network. But instead of directly defining on the weight matrices, we define it on the \update{parameterized per-layer Lipschitz bounds $\softplus(c_i)$} in the normalization layer \refequ{normalization_layer}. Specifically, we augment the original loss function $\mathcal{L}$ with a Lipschitz term as 
%
\begin{align}\label{equ:add_regularization}
  \update{\boxed{\mathcal{J}(\theta, C) = \mathcal{L}(\theta) + \alpha ∏_{i = 1}^l \softplus(c_i)}}
\end{align}
where we use $C = \{ c_i \}$ to denote the collection of per-layer Lipschitz constants $c_i$ used in the weight normalization. As mentioned in \refequ{lipschitz_bound_MLP}, the product of \update{per-layer Lipschitz constants} is the Lipschitz bound of the network. 

\subsection{Comparison with Alternatives}\label{sec:comparisons}
There are many ways one can implement and formulate a Lipschitz regularization. In this section, we compare our formulation with alternative formulations. As the amount of regularization $\alpha$ will influence the analysis, we perform parameter sweeping for each formulation independently and compare their best set-ups. 

\begin{figure}
  \begin{center}
  \includegraphics[width=1\linewidth]{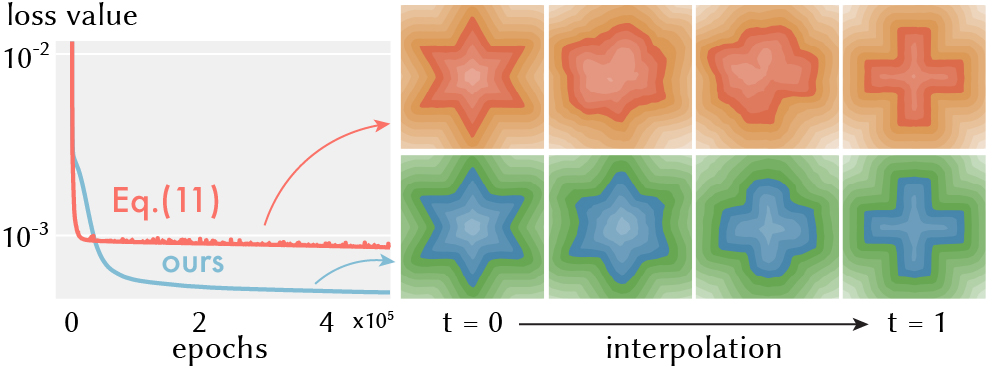}
  \end{center}
  \vspace{-5pt}
  \caption{Our method converges to a smoother result compared to the $k$-Lipschitz architecture described in \cite{AnilLG19} (see \refequ{james}). We use the same $\alpha$ for both networks because we both define the regularization as the raw Lipschitz constant of the network. }
  \label{fig:james}
  \vspace{-5pt}
\end{figure}
One solution is to design a regularization based on the architecture of the $k$-Lipschitz networks, such as the one suggested by \citet{AnilLG19}. Specifically, \citet{AnilLG19} constrain all the layers to be $1$-Lipschitz and multiply the final layer with a constant $k$ to make it $k$-Lipschitz. A possible formulation to make it learnable is to simply treat the $k$ as the Lipschitz regularization term
\begin{align}\label{equ:james}
  \mathcal{J}(\theta, k) = \mathcal{L}(\theta) + \alpha k
\end{align}
%
%
%
However, we struggle to use this formulation in \refequ{james} to find a good local minimum even for the simple 2D interpolation task (see \reffig{james}).
Moreover, when one switches to other types of activations, the result is even worse because the distributive property of per-layer scaling no longer holds.

%
\update{Another alternative is the formulation} by \citet{YoshidaM17} which defines a Lipschitz-like regularization as the summation of squared Lipschitz bounds of each layer. \update{Generalizing the definition in \cite{YoshidaM17} to $p$-norms gives us}
\begin{align}\label{equ:yoshida}
  \mathcal{J}(\theta) = \mathcal{L}(\theta) + \alpha ∑_{i = 1}^l \| \mW_i \|_p^2
\end{align}
Although this formulation can effectively find a smooth solution given a good $\alpha$, finding a good $\alpha$ is not easy with this formulation. This formulation fails to capture the exponential growth of the Lipschitz constant with respect to the network depth (see \refequ{lipschitz_bound_MLP}). In practice, it implies that the formulation \refequ{yoshida} proposed by \citet{YoshidaM17} requires a different $\alpha$ when we change the depth of the network. In \reffig{yoshida_compare}, we use the spectral norm for the method by \citet{YoshidaM17} and show that the same $\alpha$ results in different behavior for networks with different capacities. In contrast, our method leads to a more consistent behavior with the same $\alpha$.

Another alternative is to define the Lipschitz regularization directly on the weight matrices, without the normalization layer.
\begin{align}\label{equ:direct_weight}
  \mathcal{J}(\theta) = \mathcal{L}(\theta) + \alpha ∏_{i = 1}^l \| \mW_i \|_\infty
\end{align}
\begin{wrapfigure}[8]{r}{0.4\linewidth}
  \centering
  \vspace*{-1.1\intextsep}
  \hspace*{-.75\columnsep}
  \includegraphics[width=1.15\linewidth, trim={0mm 2mm 0mm 0mm}]{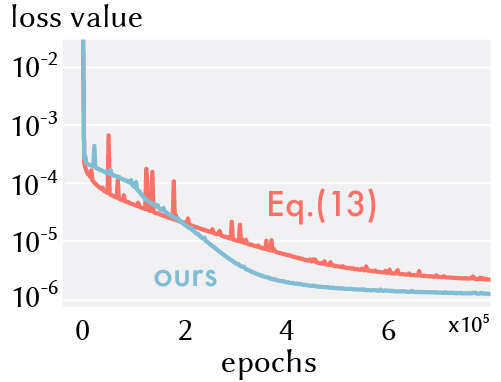}
\end{wrapfigure}
\update{This approach works equally well as our method on narrower networks, but it converges slower on wider ones}. We suspect that this is because the $\infty$-norm only depends on a single row of the weight matrix. So on a wider network, \update{this formulation requires more epochs to penalize its parameters}. In the inset, \update{we show the convergence of a 2-layer MLP with 1024 neurons each layer on 2D interpolation.}

Another tempting solution is to consider the log of the Lipschitz bound to turn the product in \refequ{add_regularization} into a summation 
\begin{align}\label{equ:log_lipschitz}
  \update{\mathcal{J}(\theta, C) = \mathcal{L}(\theta) + \alpha \sum_{i = 1}^l \log(\softplus(c_i))}
\end{align}
\begin{wrapfigure}[8]{r}{0.4\linewidth}
  \centering
  \vspace*{-0.9\intextsep}
  \hspace*{-.75\columnsep}
  \includegraphics[width=1.15\linewidth, trim={0mm 2mm 0mm 0mm}]{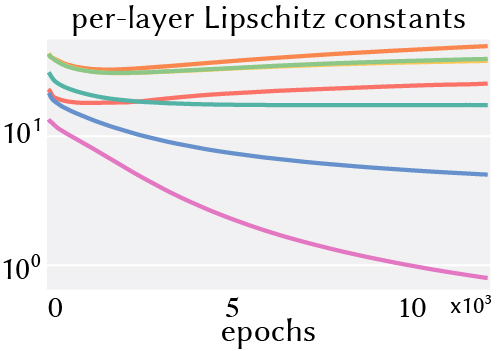}
\end{wrapfigure}
However, this makes the regularization unbounded because log goes to negative infinity when one of the Lipschitz constants approaches zero. In practice, this implies the tendency to continue penalizing the layer with a smaller Lipschitz constant. In a few cases, we did not observe the \update{Lipschitz constants} to converge. In the inset, we visualize the per-layer Lipschitz constants of a network trained on the ShapeNet \cite{shapenet}, where the layers are ordered in rainbow colors. \update{We can observe that one of the Lipschitz constants continues to decrease (purple) even after a week of training.}

%
\begin{figure*}
  \includegraphics[width=\linewidth]{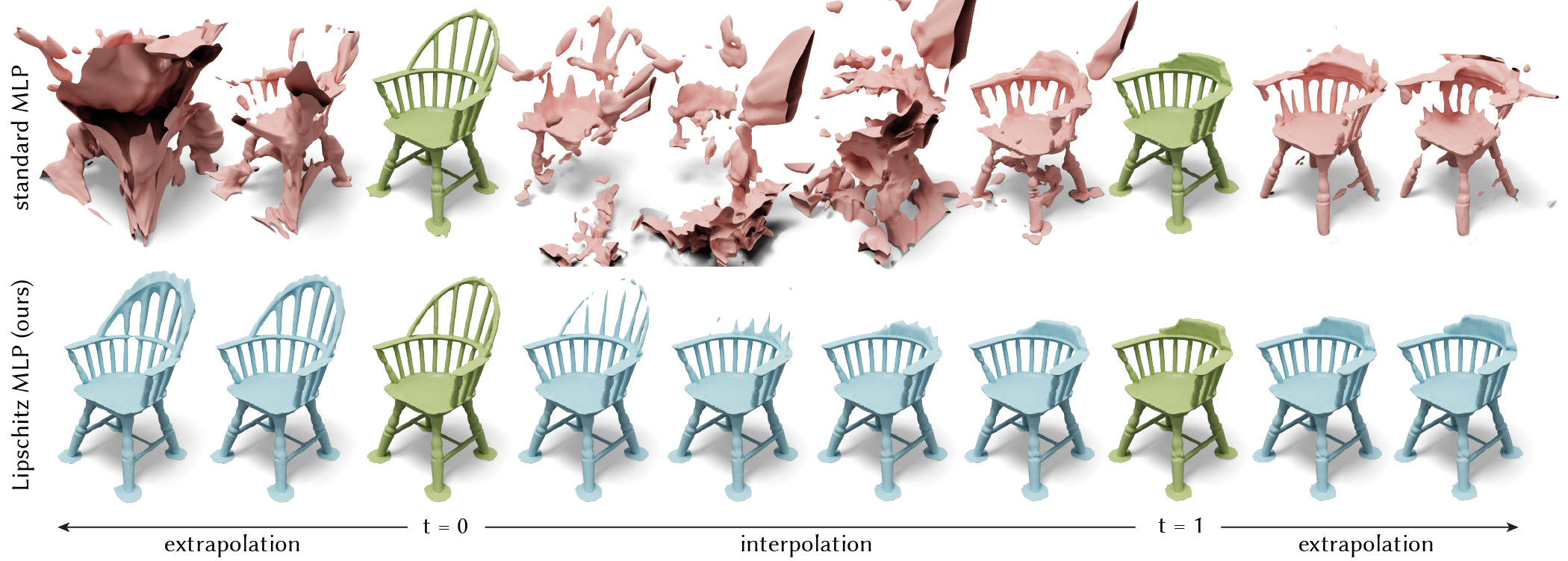}
  \vspace{-5pt}
  \caption{Our Lipschitz regularization is applicable to different implicit representations. We \update{fit} an occupancy network \cite{MeschederONNGIGM19} to two shapes (green) at $t=0$ and $t=1$ with (blue) and without (red) our Lipschitz regularization. \update{Our method generates smoother interpolation/extrapolation results.}}
  \label{fig:occupancy_interp} 
\end{figure*}
\begin{table}
  \setlength{\tabcolsep}{5.425pt}
  \centering
  \caption{
    We compute the squared norm of the Jacobian matrix $\mJ$ via back-propagation. We report the average and the maximum value of the $\| \mJ \|^2$ for all training data and show that our Lipschitz regularization achieves a smoother solution compared to other weight decay methods.
  }
  \vspace{-5pt}
  \begin{tabularx}{0.755\linewidth}{l|ccccc}
      \toprule
      \textit{Metrics} & \text{Ours} & \text{L2} & \text{L1} & \text{Vanilla}\\
      \rowcolor{derekTableBlue}
      \text{mean $\| \mJ \|^2$}  & \textbf{1.009} & 1.020 & 1.016 & 1.021 \\
      \text{max $\| \mJ \|^2$} & \textbf{9.419} & 21.181 & 17.361 & 23.658 \\
      \bottomrule
  \end{tabularx}
  \label{tab:smoothness_quantitative}
\end{table}
\begin{figure}
  \begin{center}
  \includegraphics[width=1\linewidth]{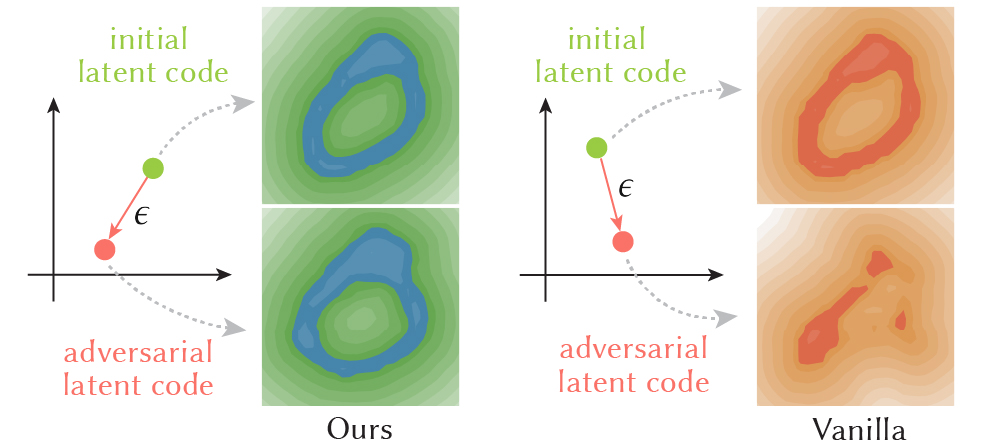}
  \end{center}
  \vspace{-5pt}
  \caption{We perform a (virtual) adversarial attack which perturbs the latent code with the fast gradient sign method \cite{GoodfellowSS14}. Vanilla AE is vulnerable to the attack so the SDF of ``0'' is completely destroyed. In contrast, our Lipschitz regularized network is more robust \update{to} the attack.}
  \label{fig:virtual_adversarial}
  \vspace{-5pt}
\end{figure}
\subsection{Comparison with Weight Decay}\label{sec:weight_decay}
Our Lipschitz regularization can be perceived as a variant of the weight decay regularization, such as the Tikhonov (L2) regularization \cite{tihonov1963solution} and Lasso (L1) \cite{tibshirani1996regression}. 
These weight decay methods are often used to avoid overfitting and improve generalization ability. However, it is unclear what their relationships are with respect to the smoothness of the network. As a result, 
\begin{wrapfigure}[7]{r}{0.4\linewidth}
  \centering
  \vspace*{-1\intextsep}
  \hspace*{-.75\columnsep}
  \includegraphics[width=1.15\linewidth, trim={0mm 2mm 0mm 0mm}]{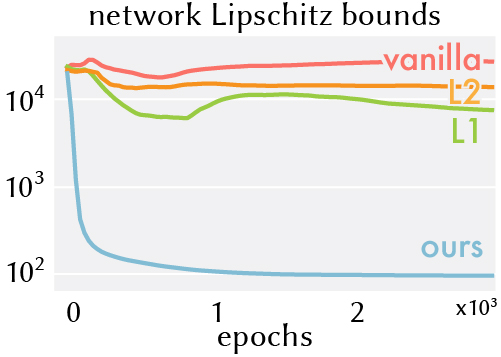}
\end{wrapfigure}
networks trained using weight decay are less smooth (measured by Lipschitz constant) compared to the network trained with our Lipschitz regularization (see the inset).

Our Lipschitz regularization also leads to a smoother network compared to other weight decay measured by a popular metric, square Jacobian norm.
To verify this, we train autoencoders (AE) to reconstruct the MNIST digits represented as signed distance functions. In \reftab{smoothness_quantitative}, our Lipschitz AE leads to smaller Jacobian norms compared to the vanilla AE, the L1 regularized AE, and the L2 regularized AE.
We provide experimental details in \refapp{MNIST}.

Besides weight decay, there are other types of regularization that are not defined on network weights, such as adding noise \cite{PooleSG14} and Dropout \cite{SrivastavaHKSS14}. These methods can complement our approach, such as \cite{GoukFPC21}. We leave the study on mixing and matching these regularizations as future work. For a more comprehensive discussion, please refer to a survey on regularization \cite{moradi2020survey}. 
\begin{figure*}
  \includegraphics[width=\linewidth, trim={0mm 5mm 0mm 0mm}]{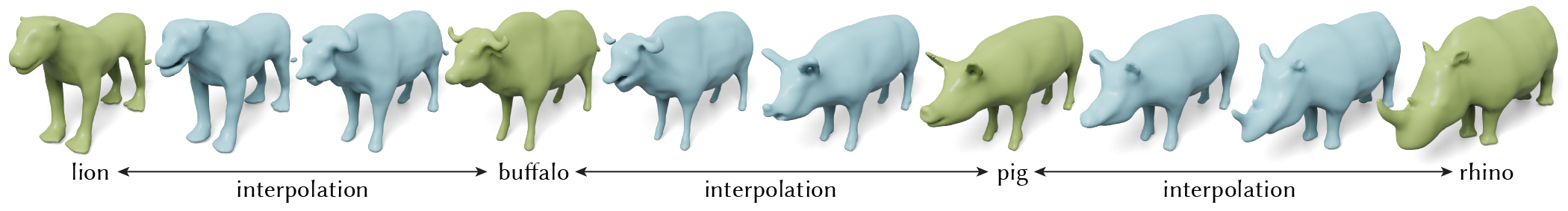}
  \vspace{1pt}
  \caption{Our method only encourages smooth interpolation. Thus our method cannot extract high-level information, such as semantics, from only a handful of shapes. Therefore, when we interpolate between animal shapes (green), the interpolated results (blue) may not be realistic animals.}
  \label{fig:multiSDF_interp} 
\end{figure*}
\section{Experiments}\label{sec:results}
Our regularization encourages a fully connected network to output Lipschitz continuous functions and is therefore applicable to different tasks that favor smooth solutions. In this section, we examine at the effectiveness of our approach in improving the robustness of a network, shape interpolation, and test-time optimization.

%
\subsection{Adversarial Robustness}\label{sec:robustness}
Adversarial attacks are small, structured changes made to a network's input signal that cause a significant change in output \cite{SzegedyZSBEGF13}. As been previously shown, Lipschitz continuous networks can improve robustness against adversarial attacks \cite{LiHALGJ19}. Here we demonstrate that our proposed regularization can serve that purpose. 
To that end we train an AE to reconstruct the signed distance functions of MNIST digits from their input image. We then adversarially perturb the latent code as described in \reffig{virtual_adversarial}, and show that Lipschitz MLP is more robust to adversarial perturbations than a standard one. We quantitatively evaluate the robustness against this type of latent adversarial attack on all the MNIST digits. A standard AE results in an average $0.06$ and maximum $0.34$ difference in the signed distance value. In contrast, our Lipschitz AE is more robust with only an average $0.03$ and maximum $0.16$ difference. 
We refer readers to \refapp{MNIST} for details about this experiment. 

\subsection{Few-Shot Shape Interpolation \& Extrapolation}\label{sec:interpolation_extrapolation}
Shape interpolation is a fundamental task and several classic methods exist, such as \cite{SolomonGPCBNDG15} and \cite{KilianMP07}. 
When shapes are mapped from a latent descriptor to 3D via a decoder, interpolation through latent space traversal often requires training on abundant data so that the latent space is well structured. Our regularization can aid shape interpolation and extrapolation in given only sparse training shapes. \updateNew{In \reffig{more_sdf_interp}, we provide several 3D SDF interpolation examples trained on only two shapes.}
%
%
Our method is also applicable to other implicit representations. In \reffig{occupancy_interp}, we evaluate our method to interpolate the \emph{occupancy} \cite{MeschederONNGIGM19} which assigns each point in $\R^3$ a binary value $[0,1]$ representing whether it is inside or outside.
\begin{figure}
  \begin{center}
  \includegraphics[width=1\linewidth, trim={0mm 2mm 0mm 0mm}]{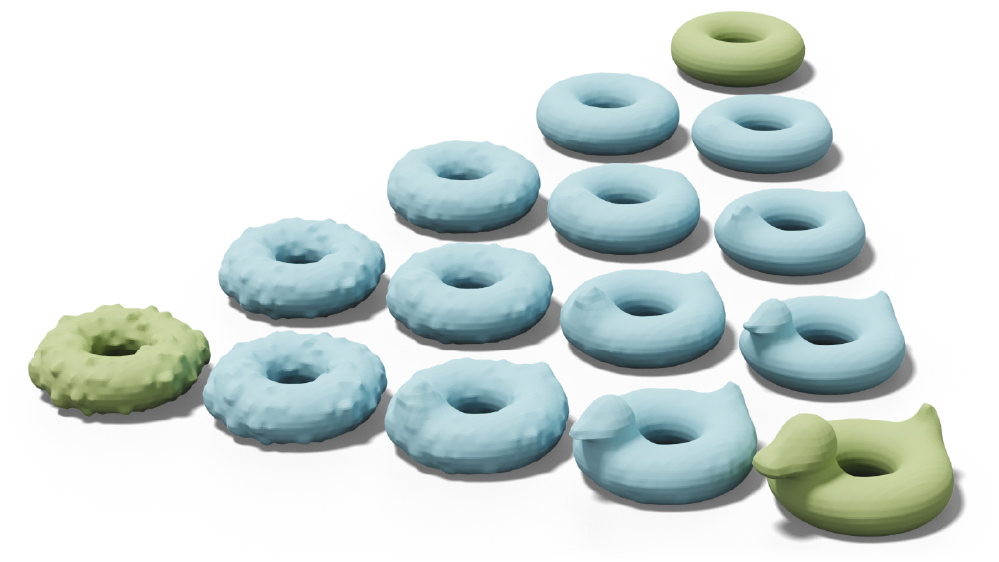}
  \end{center}
  \caption{Our method enables smooth interpolation between few training shapes. By training our model on three examples (green), we can generate high-quality novel shapes by interpolating latent codes of training shapes.}
  \label{fig:3d_interpolation}
\end{figure}

\subsection{Reconstruction with Test Time Optimization}\label{sec:testtime}
Autoencoders are popular in reconstructing the full shape from a partial point cloud. However, simply forward passing the partial points through the AE often outputs unsatisfying results. 
A common way to resolve this issue is to further optimize the latent code of the partial point cloud during test time \cite{GurumurthyA19}. Despite being effective, test time optimization is very sensitive to parameters in the optimization (e.g., initialization) and suffers from bad local minima. \citet{duggal2022mending} even propose a dedicated method aiming for resolving this issue of test-time optimization. 

\begin{figure}
  \begin{center}
  \includegraphics[width=1\linewidth, trim={0mm 2mm 0mm 0mm}]{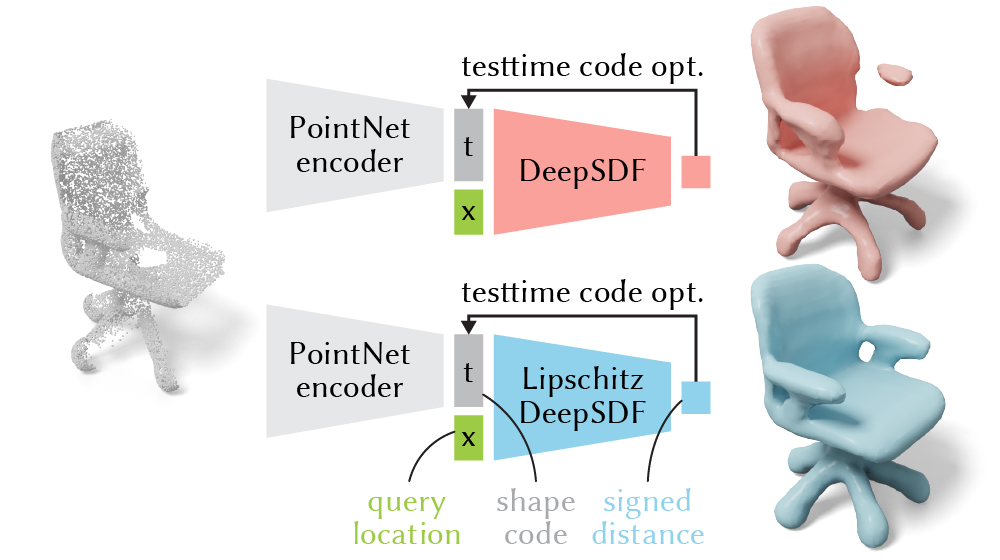}
  \end{center} 
  \caption{Encouraging smoothness with respect to the change in latent code encourages the latent space to be more compact. In the application of test time optimization, our method yields a better reconstruction (blue) given partial observations (left).}
  \label{fig:testtime_qualitative} 
\end{figure}

We discover that our Lipschitz regularization can complement the research in stabilizing test time optimization. Simply by adding our Lipschitz regularization to the vanilla autoencoder set-up, we can encourage a smoother latent manifold and stabilize the test time optimization. In \reffig{testtime_qualitative} and \reftab{testtime_quantitative}, we show that we achieve a better reconstruction result both qualitatively and quantitatively. 
Our method can complement the method based on training additional networks, such as adding a discriminator \cite{duggal2022mending} or a generative adversarial network \cite{GurumurthyA19}. But we leave them as future work.
\begin{table}
  \setlength{\tabcolsep}{5.425pt}
  \centering
  \caption{
    We quantitatively evaluate the test time optimization. Given a ground truth point cloud from the test set, we delete the right-half of the point to obtain a partial point cloud and then we perform test-time optimization to reconstruct the full shape back. We report the Chamfer distance and Hausdorff distance between the ground truth point cloud and our reconstructed full shape averaged across the test set.}
  \begin{tabularx}{0.975\linewidth}{l|cc}
      \toprule
      \textit{Metrics} & Lipschitz DeepSDF & DeepSDF \\
      \rowcolor{derekTableBlue}
      \text{average Chamfer distance} & \textbf{0.0013} & 0.0343 \\
      \text{average Hausdorff distance} & \textbf{0.1270} & 0.3441\\
      \bottomrule
  \end{tabularx}
  \label{tab:testtime_quantitative} 
\end{table}

\section{Conclusion \& Future Work}\label{sec:conclusion}
Our regularization encourages fully connected networks to have a small Lipschitz constant.
%
%
Our regularization is defined on a loose upper bound of the true Lipschitz constant. Using a tighter estimate would benefit applications that require more precise control. 
Our shape interpolation behaves similarly to linear interpolation. Incorporating the Wasserstein metric into our smoothness measure could encourage shape interpolation to behave more like \emph{optimal transport}. 
Furthermore, encouraging learning high-level structural information from few examples would aid to the experiment on few shot shape interpolation (see \reffig{multiSDF_interp}). 
Our Lipschitz regularization is a generic technique to encourage smooth neural network solutions. However, our experiments are mainly conducted on neural implicit geometry tasks. We would be interested in applying our regularization to other tasks beyond geometry processing.

\begin{acks}
    Our research is funded in part by NSERC Discovery (RGPIN2017–05235, RGPAS–2017–507938), New Frontiers of Research Fund (NFRFE–201), the Ontario Early Research Award program, the Canada Research Chairs Program, a Sloan Research Fellowship, the DSI Catalyst Grant program and gifts by Adobe Systems.
    Special thanks to Otman Benchekroun, Alex Evans, James Lucas, Oded Stein, Towaki Takikawa, and Jackson Wang for inspiring early discussions. We thank Silvia Sell\'{a}n and Frank Shen for assisting code release and experiments. We also want to thank all the artists for sharing a rich variety of 3D models to facilitate our research.
\end{acks}


\bibliographystyle{ACM-Reference-Format}
\bibliography{sections/references}


\begin{thebibliography}{56}


\ifx \showCODEN    \undefined \def \showCODEN     #1{\unskip}     \fi
\ifx \showDOI      \undefined \def \showDOI       #1{#1}\fi
\ifx \showISBNx    \undefined \def \showISBNx     #1{\unskip}     \fi
\ifx \showISBNxiii \undefined \def \showISBNxiii  #1{\unskip}     \fi
\ifx \showISSN     \undefined \def \showISSN      #1{\unskip}     \fi
\ifx \showLCCN     \undefined \def \showLCCN      #1{\unskip}     \fi
\ifx \shownote     \undefined \def \shownote      #1{#1}          \fi
\ifx \showarticletitle \undefined \def \showarticletitle #1{#1}   \fi
\ifx \showURL      \undefined \def \showURL       {\relax}        \fi
\providecommand\bibfield[2]{#2}
\providecommand\bibinfo[2]{#2}
\providecommand\natexlab[1]{#1}
\providecommand\showeprint[2][]{arXiv:#2}

\bibitem[Anil et~al\mbox{.}(2019)]%
        {AnilLG19}
\bibfield{author}{\bibinfo{person}{Cem Anil}, \bibinfo{person}{James Lucas},
  {and} \bibinfo{person}{Roger~B. Grosse}.} \bibinfo{year}{2019}\natexlab{}.
\newblock \showarticletitle{Sorting Out Lipschitz Function Approximation}. In
  \bibinfo{booktitle}{\emph{Proceedings of the 36th International Conference on
  Machine Learning, {ICML} 2019, 9-15 June 2019, Long Beach, California,
  {USA}}} \emph{(\bibinfo{series}{Proceedings of Machine Learning Research},
  Vol.~\bibinfo{volume}{97})}, \bibfield{editor}{\bibinfo{person}{Kamalika
  Chaudhuri} {and} \bibinfo{person}{Ruslan Salakhutdinov}} (Eds.).
  \bibinfo{publisher}{{PMLR}}, \bibinfo{pages}{291--301}.
\newblock


\bibitem[Arjovsky et~al\mbox{.}(2017)]%
        {arjovsky2017wasserstein}
\bibfield{author}{\bibinfo{person}{Martin Arjovsky}, \bibinfo{person}{Soumith
  Chintala}, {and} \bibinfo{person}{L{\'e}on Bottou}.}
  \bibinfo{year}{2017}\natexlab{}.
\newblock \showarticletitle{Wasserstein generative adversarial networks}. In
  \bibinfo{booktitle}{\emph{International conference on machine learning}}.
  PMLR, \bibinfo{pages}{214--223}.
\newblock


\bibitem[Bj{\"o}rck and Bowie(1971)]%
        {bjorck1971iterative}
\bibfield{author}{\bibinfo{person}{{\AA}ke Bj{\"o}rck} {and}
  \bibinfo{person}{Clazett Bowie}.} \bibinfo{year}{1971}\natexlab{}.
\newblock \showarticletitle{An iterative algorithm for computing the best
  estimate of an orthogonal matrix}.
\newblock \bibinfo{journal}{\emph{SIAM J. Numer. Anal.}} \bibinfo{volume}{8},
  \bibinfo{number}{2} (\bibinfo{year}{1971}), \bibinfo{pages}{358--364}.
\newblock


\bibitem[Bradbury et~al\mbox{.}(2018)]%
        {jax2018github}
\bibfield{author}{\bibinfo{person}{James Bradbury}, \bibinfo{person}{Roy
  Frostig}, \bibinfo{person}{Peter Hawkins}, \bibinfo{person}{Matthew~James
  Johnson}, \bibinfo{person}{Chris Leary}, \bibinfo{person}{Dougal Maclaurin},
  \bibinfo{person}{George Necula}, \bibinfo{person}{Adam Paszke},
  \bibinfo{person}{Jake Vander{P}las}, \bibinfo{person}{Skye
  Wanderman-{M}ilne}, {and} \bibinfo{person}{Qiao Zhang}.}
  \bibinfo{year}{2018}\natexlab{}.
\newblock \bibinfo{booktitle}{\emph{{JAX}: composable transformations of
  {P}ython+{N}um{P}y programs}}.
\newblock
\urldef\tempurl%
\url{http://github.com/google/jax}
\showURL{%
\tempurl}


\bibitem[Chang et~al\mbox{.}(2015)]%
        {shapenet}
\bibfield{author}{\bibinfo{person}{Angel~X. Chang}, \bibinfo{person}{Thomas~A.
  Funkhouser}, \bibinfo{person}{Leonidas~J. Guibas}, \bibinfo{person}{Pat
  Hanrahan}, \bibinfo{person}{Qi{-}Xing Huang}, \bibinfo{person}{Zimo Li},
  \bibinfo{person}{Silvio Savarese}, \bibinfo{person}{Manolis Savva},
  \bibinfo{person}{Shuran Song}, \bibinfo{person}{Hao Su},
  \bibinfo{person}{Jianxiong Xiao}, \bibinfo{person}{Li Yi}, {and}
  \bibinfo{person}{Fisher Yu}.} \bibinfo{year}{2015}\natexlab{}.
\newblock \showarticletitle{ShapeNet: An Information-Rich 3D Model Repository}.
\newblock \bibinfo{journal}{\emph{CoRR}}  \bibinfo{volume}{abs/1512.03012}
  (\bibinfo{year}{2015}).
\newblock


\bibitem[Chen et~al\mbox{.}(2019)]%
        {ChenLGSLJF19}
\bibfield{author}{\bibinfo{person}{Wenzheng Chen}, \bibinfo{person}{Huan Ling},
  \bibinfo{person}{Jun Gao}, \bibinfo{person}{Edward~J. Smith},
  \bibinfo{person}{Jaakko Lehtinen}, \bibinfo{person}{Alec Jacobson}, {and}
  \bibinfo{person}{Sanja Fidler}.} \bibinfo{year}{2019}\natexlab{}.
\newblock \showarticletitle{Learning to Predict 3D Objects with an
  Interpolation-based Differentiable Renderer}. In
  \bibinfo{booktitle}{\emph{Advances in Neural Information Processing Systems
  32: Annual Conference on Neural Information Processing Systems 2019, NeurIPS
  2019, December 8-14, 2019, Vancouver, BC, Canada}},
  \bibfield{editor}{\bibinfo{person}{Hanna~M. Wallach}, \bibinfo{person}{Hugo
  Larochelle}, \bibinfo{person}{Alina Beygelzimer}, \bibinfo{person}{Florence
  d'Alch{\'{e}}{-}Buc}, \bibinfo{person}{Emily~B. Fox}, {and}
  \bibinfo{person}{Roman Garnett}} (Eds.). \bibinfo{pages}{9605--9616}.
\newblock


\bibitem[Ciss{\'{e}} et~al\mbox{.}(2017)]%
        {CisseBGDU17}
\bibfield{author}{\bibinfo{person}{Moustapha Ciss{\'{e}}},
  \bibinfo{person}{Piotr Bojanowski}, \bibinfo{person}{Edouard Grave},
  \bibinfo{person}{Yann~N. Dauphin}, {and} \bibinfo{person}{Nicolas Usunier}.}
  \bibinfo{year}{2017}\natexlab{}.
\newblock \showarticletitle{Parseval Networks: Improving Robustness to
  Adversarial Examples}. In \bibinfo{booktitle}{\emph{Proceedings of the 34th
  International Conference on Machine Learning, {ICML} 2017, Sydney, NSW,
  Australia, 6-11 August 2017}} \emph{(\bibinfo{series}{Proceedings of Machine
  Learning Research}, Vol.~\bibinfo{volume}{70})},
  \bibfield{editor}{\bibinfo{person}{Doina Precup} {and}
  \bibinfo{person}{Yee~Whye Teh}} (Eds.). \bibinfo{publisher}{{PMLR}},
  \bibinfo{pages}{854--863}.
\newblock


\bibitem[Drucker and Le~Cun(1991)]%
        {drucker1991double}
\bibfield{author}{\bibinfo{person}{Harris Drucker} {and} \bibinfo{person}{Yann
  Le~Cun}.} \bibinfo{year}{1991}\natexlab{}.
\newblock \showarticletitle{Double backpropagation increasing generalization
  performance}. In \bibinfo{booktitle}{\emph{IJCNN-91-Seattle International
  Joint Conference on Neural Networks}}, Vol.~\bibinfo{volume}{2}. IEEE,
  \bibinfo{pages}{145--150}.
\newblock


\bibitem[Duggal et~al\mbox{.}(2022)]%
        {duggal2022mending}
\bibfield{author}{\bibinfo{person}{Shivam Duggal}, \bibinfo{person}{Zihao
  Wang}, \bibinfo{person}{Wei-Chiu Ma}, \bibinfo{person}{Sivabalan
  Manivasagam}, \bibinfo{person}{Justin Liang}, \bibinfo{person}{Shenlong
  Wang}, {and} \bibinfo{person}{Raquel Urtasun}.}
  \bibinfo{year}{2022}\natexlab{}.
\newblock \showarticletitle{Mending Neural Implicit Modeling for 3D Vehicle
  Reconstruction in the Wild}. In \bibinfo{booktitle}{\emph{Proceedings of the
  IEEE/CVF Winter Conference on Applications of Computer Vision}}.
  \bibinfo{pages}{1900--1909}.
\newblock


\bibitem[Elsner et~al\mbox{.}(2021)]%
        {Elsner2021Lipschitz}
\bibfield{author}{\bibinfo{person}{Tim Elsner}, \bibinfo{person}{Moritz Ibing},
  \bibinfo{person}{Victor Czech}, \bibinfo{person}{Julius Nehring{-}Wirxel},
  {and} \bibinfo{person}{Leif Kobbelt}.} \bibinfo{year}{2021}\natexlab{}.
\newblock \showarticletitle{Intuitive Shape Editing in Latent Space}.
\newblock \bibinfo{journal}{\emph{CoRR}}  \bibinfo{volume}{abs/2111.12488}
  (\bibinfo{year}{2021}).
\newblock
\showeprint[arXiv]{2111.12488}


\bibitem[Glorot and Bengio(2010)]%
        {GlorotB10}
\bibfield{author}{\bibinfo{person}{Xavier Glorot} {and} \bibinfo{person}{Yoshua
  Bengio}.} \bibinfo{year}{2010}\natexlab{}.
\newblock \showarticletitle{Understanding the difficulty of training deep
  feedforward neural networks}. In \bibinfo{booktitle}{\emph{Proceedings of the
  Thirteenth International Conference on Artificial Intelligence and
  Statistics, {AISTATS} 2010, Chia Laguna Resort, Sardinia, Italy, May 13-15,
  2010}} \emph{(\bibinfo{series}{{JMLR} Proceedings},
  Vol.~\bibinfo{volume}{9})}, \bibfield{editor}{\bibinfo{person}{Yee~Whye Teh}
  {and} \bibinfo{person}{D.~Mike Titterington}} (Eds.).
  \bibinfo{publisher}{JMLR.org}, \bibinfo{pages}{249--256}.
\newblock


\bibitem[Goodfellow et~al\mbox{.}(2015)]%
        {GoodfellowSS14}
\bibfield{author}{\bibinfo{person}{Ian~J. Goodfellow},
  \bibinfo{person}{Jonathon Shlens}, {and} \bibinfo{person}{Christian
  Szegedy}.} \bibinfo{year}{2015}\natexlab{}.
\newblock \showarticletitle{Explaining and Harnessing Adversarial Examples}. In
  \bibinfo{booktitle}{\emph{3rd International Conference on Learning
  Representations, {ICLR} 2015, San Diego, CA, USA, May 7-9, 2015, Conference
  Track Proceedings}}, \bibfield{editor}{\bibinfo{person}{Yoshua Bengio} {and}
  \bibinfo{person}{Yann LeCun}} (Eds.).
\newblock


\bibitem[Gouk et~al\mbox{.}(2021)]%
        {GoukFPC21}
\bibfield{author}{\bibinfo{person}{Henry Gouk}, \bibinfo{person}{Eibe Frank},
  \bibinfo{person}{Bernhard Pfahringer}, {and} \bibinfo{person}{Michael~J.
  Cree}.} \bibinfo{year}{2021}\natexlab{}.
\newblock \showarticletitle{Regularisation of neural networks by enforcing
  Lipschitz continuity}.
\newblock \bibinfo{journal}{\emph{Mach. Learn.}} \bibinfo{volume}{110},
  \bibinfo{number}{2} (\bibinfo{year}{2021}), \bibinfo{pages}{393--416}.
\newblock


\bibitem[Gropp et~al\mbox{.}(2020)]%
        {GroppYHAL20}
\bibfield{author}{\bibinfo{person}{Amos Gropp}, \bibinfo{person}{Lior Yariv},
  \bibinfo{person}{Niv Haim}, \bibinfo{person}{Matan Atzmon}, {and}
  \bibinfo{person}{Yaron Lipman}.} \bibinfo{year}{2020}\natexlab{}.
\newblock \showarticletitle{Implicit Geometric Regularization for Learning
  Shapes}. In \bibinfo{booktitle}{\emph{Proceedings of the 37th International
  Conference on Machine Learning, {ICML} 2020, 13-18 July 2020, Virtual Event}}
  \emph{(\bibinfo{series}{Proceedings of Machine Learning Research},
  Vol.~\bibinfo{volume}{119})}. \bibinfo{publisher}{{PMLR}},
  \bibinfo{pages}{3789--3799}.
\newblock


\bibitem[Gulrajani et~al\mbox{.}(2017)]%
        {GulrajaniAADC17}
\bibfield{author}{\bibinfo{person}{Ishaan Gulrajani}, \bibinfo{person}{Faruk
  Ahmed}, \bibinfo{person}{Mart{\'{\i}}n Arjovsky}, \bibinfo{person}{Vincent
  Dumoulin}, {and} \bibinfo{person}{Aaron~C. Courville}.}
  \bibinfo{year}{2017}\natexlab{}.
\newblock \showarticletitle{Improved Training of Wasserstein GANs}. In
  \bibinfo{booktitle}{\emph{Advances in Neural Information Processing Systems
  30: Annual Conference on Neural Information Processing Systems 2017, December
  4-9, 2017, Long Beach, CA, {USA}}},
  \bibfield{editor}{\bibinfo{person}{Isabelle Guyon}, \bibinfo{person}{Ulrike
  von Luxburg}, \bibinfo{person}{Samy Bengio}, \bibinfo{person}{Hanna~M.
  Wallach}, \bibinfo{person}{Rob Fergus}, \bibinfo{person}{S.~V.~N.
  Vishwanathan}, {and} \bibinfo{person}{Roman Garnett}} (Eds.).
  \bibinfo{pages}{5767--5777}.
\newblock


\bibitem[Gurumurthy and Agrawal(2019)]%
        {GurumurthyA19}
\bibfield{author}{\bibinfo{person}{Swaminathan Gurumurthy} {and}
  \bibinfo{person}{Shubham Agrawal}.} \bibinfo{year}{2019}\natexlab{}.
\newblock \showarticletitle{High Fidelity Semantic Shape Completion for Point
  Clouds Using Latent Optimization}. In \bibinfo{booktitle}{\emph{{IEEE} Winter
  Conference on Applications of Computer Vision, {WACV} 2019, Waikoloa Village,
  HI, USA, January 7-11, 2019}}. \bibinfo{publisher}{{IEEE}},
  \bibinfo{pages}{1099--1108}.
\newblock


\bibitem[He et~al\mbox{.}(2015)]%
        {HeZRS15}
\bibfield{author}{\bibinfo{person}{Kaiming He}, \bibinfo{person}{Xiangyu
  Zhang}, \bibinfo{person}{Shaoqing Ren}, {and} \bibinfo{person}{Jian Sun}.}
  \bibinfo{year}{2015}\natexlab{}.
\newblock \showarticletitle{Delving Deep into Rectifiers: Surpassing
  Human-Level Performance on ImageNet Classification}. In
  \bibinfo{booktitle}{\emph{2015 {IEEE} International Conference on Computer
  Vision, {ICCV} 2015, Santiago, Chile, December 7-13, 2015}}.
  \bibinfo{publisher}{{IEEE} Computer Society}, \bibinfo{pages}{1026--1034}.
\newblock


\bibitem[Hertz et~al\mbox{.}(2020)]%
        {HertzHGC20}
\bibfield{author}{\bibinfo{person}{Amir Hertz}, \bibinfo{person}{Rana Hanocka},
  \bibinfo{person}{Raja Giryes}, {and} \bibinfo{person}{Daniel Cohen{-}Or}.}
  \bibinfo{year}{2020}\natexlab{}.
\newblock \showarticletitle{Deep geometric texture synthesis}.
\newblock \bibinfo{journal}{\emph{{ACM} Trans. Graph.}} \bibinfo{volume}{39},
  \bibinfo{number}{4} (\bibinfo{year}{2020}), \bibinfo{pages}{108}.
\newblock


\bibitem[Hoffman et~al\mbox{.}(2019)]%
        {judy2019jacobian}
\bibfield{author}{\bibinfo{person}{Judy Hoffman}, \bibinfo{person}{Daniel~A.
  Roberts}, {and} \bibinfo{person}{Sho Yaida}.}
  \bibinfo{year}{2019}\natexlab{}.
\newblock \showarticletitle{Robust Learning with Jacobian Regularization}.
\newblock \bibinfo{journal}{\emph{CoRR}}  \bibinfo{volume}{abs/1908.02729}
  (\bibinfo{year}{2019}).
\newblock
\showeprint[arXiv]{1908.02729}


\bibitem[Horn and Johnson(2012)]%
        {horn2012matrix}
\bibfield{author}{\bibinfo{person}{Roger~A Horn} {and}
  \bibinfo{person}{Charles~R Johnson}.} \bibinfo{year}{2012}\natexlab{}.
\newblock \bibinfo{booktitle}{\emph{Matrix analysis}}.
\newblock \bibinfo{publisher}{Cambridge university press}.
\newblock


\bibitem[Huang et~al\mbox{.}(2018)]%
        {HuangLLYWL18}
\bibfield{author}{\bibinfo{person}{Lei Huang}, \bibinfo{person}{Xianglong Liu},
  \bibinfo{person}{Bo Lang}, \bibinfo{person}{Adams~Wei Yu},
  \bibinfo{person}{Yongliang Wang}, {and} \bibinfo{person}{Bo Li}.}
  \bibinfo{year}{2018}\natexlab{}.
\newblock \showarticletitle{Orthogonal Weight Normalization: Solution to
  Optimization Over Multiple Dependent Stiefel Manifolds in Deep Neural
  Networks}. In \bibinfo{booktitle}{\emph{Proceedings of the Thirty-Second
  {AAAI} Conference on Artificial Intelligence, (AAAI-18), the 30th innovative
  Applications of Artificial Intelligence (IAAI-18), and the 8th {AAAI}
  Symposium on Educational Advances in Artificial Intelligence (EAAI-18), New
  Orleans, Louisiana, USA, February 2-7, 2018}},
  \bibfield{editor}{\bibinfo{person}{Sheila~A. McIlraith} {and}
  \bibinfo{person}{Kilian~Q. Weinberger}} (Eds.). \bibinfo{publisher}{{AAAI}
  Press}, \bibinfo{pages}{3271--3278}.
\newblock


\bibitem[Jakubovitz and Giryes(2018)]%
        {JakubovitzG18}
\bibfield{author}{\bibinfo{person}{Daniel Jakubovitz} {and}
  \bibinfo{person}{Raja Giryes}.} \bibinfo{year}{2018}\natexlab{}.
\newblock \showarticletitle{Improving {DNN} Robustness to Adversarial Attacks
  Using Jacobian Regularization}. In \bibinfo{booktitle}{\emph{Computer Vision
  - {ECCV} 2018 - 15th European Conference, Munich, Germany, September 8-14,
  2018, Proceedings, Part {XII}}} \emph{(\bibinfo{series}{Lecture Notes in
  Computer Science}, Vol.~\bibinfo{volume}{11216})},
  \bibfield{editor}{\bibinfo{person}{Vittorio Ferrari},
  \bibinfo{person}{Martial Hebert}, \bibinfo{person}{Cristian Sminchisescu},
  {and} \bibinfo{person}{Yair Weiss}} (Eds.). \bibinfo{publisher}{Springer},
  \bibinfo{pages}{525--541}.
\newblock


\bibitem[Jordan and Dimakis(2020)]%
        {JordanD20}
\bibfield{author}{\bibinfo{person}{Matt Jordan} {and}
  \bibinfo{person}{Alexandros~G. Dimakis}.} \bibinfo{year}{2020}\natexlab{}.
\newblock \showarticletitle{Exactly Computing the Local Lipschitz Constant of
  ReLU Networks}. In \bibinfo{booktitle}{\emph{Advances in Neural Information
  Processing Systems 33: Annual Conference on Neural Information Processing
  Systems 2020, NeurIPS 2020, December 6-12, 2020, virtual}},
  \bibfield{editor}{\bibinfo{person}{Hugo Larochelle},
  \bibinfo{person}{Marc'Aurelio Ranzato}, \bibinfo{person}{Raia Hadsell},
  \bibinfo{person}{Maria{-}Florina Balcan}, {and} \bibinfo{person}{Hsuan{-}Tien
  Lin}} (Eds.).
\newblock


\bibitem[Kato et~al\mbox{.}(2018)]%
        {KatoUH18}
\bibfield{author}{\bibinfo{person}{Hiroharu Kato}, \bibinfo{person}{Yoshitaka
  Ushiku}, {and} \bibinfo{person}{Tatsuya Harada}.}
  \bibinfo{year}{2018}\natexlab{}.
\newblock \showarticletitle{Neural 3D Mesh Renderer}. In
  \bibinfo{booktitle}{\emph{2018 {IEEE} Conference on Computer Vision and
  Pattern Recognition, {CVPR} 2018, Salt Lake City, UT, USA, June 18-22,
  2018}}. \bibinfo{publisher}{Computer Vision Foundation / {IEEE} Computer
  Society}, \bibinfo{pages}{3907--3916}.
\newblock


\bibitem[Kilian et~al\mbox{.}(2007)]%
        {KilianMP07}
\bibfield{author}{\bibinfo{person}{Martin Kilian}, \bibinfo{person}{Niloy~J.
  Mitra}, {and} \bibinfo{person}{Helmut Pottmann}.}
  \bibinfo{year}{2007}\natexlab{}.
\newblock \showarticletitle{Geometric modeling in shape space}.
\newblock \bibinfo{journal}{\emph{{ACM} Trans. Graph.}} \bibinfo{volume}{26},
  \bibinfo{number}{3} (\bibinfo{year}{2007}), \bibinfo{pages}{64}.
\newblock


\bibitem[Kingma and Ba(2015)]%
        {KingmaB14}
\bibfield{author}{\bibinfo{person}{Diederik~P. Kingma} {and}
  \bibinfo{person}{Jimmy Ba}.} \bibinfo{year}{2015}\natexlab{}.
\newblock \showarticletitle{Adam: {A} Method for Stochastic Optimization}. In
  \bibinfo{booktitle}{\emph{3rd International Conference on Learning
  Representations, {ICLR} 2015, San Diego, CA, USA, May 7-9, 2015, Conference
  Track Proceedings}}, \bibfield{editor}{\bibinfo{person}{Yoshua Bengio} {and}
  \bibinfo{person}{Yann LeCun}} (Eds.).
\newblock


\bibitem[Li et~al\mbox{.}(2019a)]%
        {LiHALGJ19}
\bibfield{author}{\bibinfo{person}{Qiyang Li}, \bibinfo{person}{Saminul Haque},
  \bibinfo{person}{Cem Anil}, \bibinfo{person}{James Lucas},
  \bibinfo{person}{Roger~B. Grosse}, {and} \bibinfo{person}{J{\"{o}}rn{-}Henrik
  Jacobsen}.} \bibinfo{year}{2019}\natexlab{a}.
\newblock \showarticletitle{Preventing Gradient Attenuation in Lipschitz
  Constrained Convolutional Networks}. In \bibinfo{booktitle}{\emph{Advances in
  Neural Information Processing Systems 32: Annual Conference on Neural
  Information Processing Systems 2019, NeurIPS 2019, December 8-14, 2019,
  Vancouver, BC, Canada}}, \bibfield{editor}{\bibinfo{person}{Hanna~M.
  Wallach}, \bibinfo{person}{Hugo Larochelle}, \bibinfo{person}{Alina
  Beygelzimer}, \bibinfo{person}{Florence d'Alch{\'{e}}{-}Buc},
  \bibinfo{person}{Emily~B. Fox}, {and} \bibinfo{person}{Roman Garnett}}
  (Eds.). \bibinfo{pages}{15364--15376}.
\newblock


\bibitem[Li et~al\mbox{.}(2019b)]%
        {LiWM19a}
\bibfield{author}{\bibinfo{person}{Yuanzhi Li}, \bibinfo{person}{Colin Wei},
  {and} \bibinfo{person}{Tengyu Ma}.} \bibinfo{year}{2019}\natexlab{b}.
\newblock \showarticletitle{Towards Explaining the Regularization Effect of
  Initial Large Learning Rate in Training Neural Networks}. In
  \bibinfo{booktitle}{\emph{Advances in Neural Information Processing Systems
  32: Annual Conference on Neural Information Processing Systems 2019, NeurIPS
  2019, December 8-14, 2019, Vancouver, BC, Canada}},
  \bibfield{editor}{\bibinfo{person}{Hanna~M. Wallach}, \bibinfo{person}{Hugo
  Larochelle}, \bibinfo{person}{Alina Beygelzimer}, \bibinfo{person}{Florence
  d'Alch{\'{e}}{-}Buc}, \bibinfo{person}{Emily~B. Fox}, {and}
  \bibinfo{person}{Roman Garnett}} (Eds.). \bibinfo{pages}{11669--11680}.
\newblock


\bibitem[Liu et~al\mbox{.}(2019)]%
        {liu2019soft}
\bibfield{author}{\bibinfo{person}{Shichen Liu}, \bibinfo{person}{Tianye Li},
  \bibinfo{person}{Weikai Chen}, {and} \bibinfo{person}{Hao Li}.}
  \bibinfo{year}{2019}\natexlab{}.
\newblock \showarticletitle{Soft rasterizer: A differentiable renderer for
  image-based 3d reasoning}. In \bibinfo{booktitle}{\emph{Proceedings of the
  IEEE/CVF International Conference on Computer Vision}}.
  \bibinfo{pages}{7708--7717}.
\newblock


\bibitem[Mescheder et~al\mbox{.}(2019a)]%
        {MeschederONNGIGM19}
\bibfield{author}{\bibinfo{person}{Lars~M. Mescheder}, \bibinfo{person}{Michael
  Oechsle}, \bibinfo{person}{Michael Niemeyer}, \bibinfo{person}{Sebastian
  Nowozin}, {and} \bibinfo{person}{Andreas Geiger}.}
  \bibinfo{year}{2019}\natexlab{a}.
\newblock \showarticletitle{Occupancy Networks: Learning 3D Reconstruction in
  Function Space}. In \bibinfo{booktitle}{\emph{{IEEE} Conference on Computer
  Vision and Pattern Recognition, {CVPR} 2019, Long Beach, CA, USA, June 16-20,
  2019}}. \bibinfo{publisher}{Computer Vision Foundation / {IEEE}},
  \bibinfo{pages}{4460--4470}.
\newblock


\bibitem[Mescheder et~al\mbox{.}(2019b)]%
        {MeschederONNG19}
\bibfield{author}{\bibinfo{person}{Lars~M. Mescheder}, \bibinfo{person}{Michael
  Oechsle}, \bibinfo{person}{Michael Niemeyer}, \bibinfo{person}{Sebastian
  Nowozin}, {and} \bibinfo{person}{Andreas Geiger}.}
  \bibinfo{year}{2019}\natexlab{b}.
\newblock \showarticletitle{Occupancy Networks: Learning 3D Reconstruction in
  Function Space}. In \bibinfo{booktitle}{\emph{{IEEE} Conference on Computer
  Vision and Pattern Recognition, {CVPR} 2019, Long Beach, CA, USA, June 16-20,
  2019}}. \bibinfo{publisher}{Computer Vision Foundation / {IEEE}},
  \bibinfo{pages}{4460--4470}.
\newblock


\bibitem[Miyato et~al\mbox{.}(2018)]%
        {MiyatoKKY18}
\bibfield{author}{\bibinfo{person}{Takeru Miyato}, \bibinfo{person}{Toshiki
  Kataoka}, \bibinfo{person}{Masanori Koyama}, {and} \bibinfo{person}{Yuichi
  Yoshida}.} \bibinfo{year}{2018}\natexlab{}.
\newblock \showarticletitle{Spectral Normalization for Generative Adversarial
  Networks}. In \bibinfo{booktitle}{\emph{6th International Conference on
  Learning Representations, {ICLR} 2018, Vancouver, BC, Canada, April 30 - May
  3, 2018, Conference Track Proceedings}}. \bibinfo{publisher}{OpenReview.net}.
\newblock


\bibitem[Moosavi{-}Dezfooli et~al\mbox{.}(2019)]%
        {MoosaviDezfooli19}
\bibfield{author}{\bibinfo{person}{Seyed{-}Mohsen Moosavi{-}Dezfooli},
  \bibinfo{person}{Alhussein Fawzi}, \bibinfo{person}{Jonathan Uesato}, {and}
  \bibinfo{person}{Pascal Frossard}.} \bibinfo{year}{2019}\natexlab{}.
\newblock \showarticletitle{Robustness via Curvature Regularization, and Vice
  Versa}. In \bibinfo{booktitle}{\emph{{IEEE} Conference on Computer Vision and
  Pattern Recognition, {CVPR} 2019, Long Beach, CA, USA, June 16-20, 2019}}.
  \bibinfo{publisher}{Computer Vision Foundation / {IEEE}},
  \bibinfo{pages}{9078--9086}.
\newblock


\bibitem[Moradi et~al\mbox{.}(2020)]%
        {moradi2020survey}
\bibfield{author}{\bibinfo{person}{Reza Moradi}, \bibinfo{person}{Reza
  Berangi}, {and} \bibinfo{person}{Behrouz Minaei}.}
  \bibinfo{year}{2020}\natexlab{}.
\newblock \showarticletitle{A survey of regularization strategies for deep
  models}.
\newblock \bibinfo{journal}{\emph{Artificial Intelligence Review}}
  \bibinfo{volume}{53}, \bibinfo{number}{6} (\bibinfo{year}{2020}),
  \bibinfo{pages}{3947--3986}.
\newblock


\bibitem[Oberman and Calder(2018)]%
        {Adam2018Lipschitz}
\bibfield{author}{\bibinfo{person}{Adam~M. Oberman} {and} \bibinfo{person}{Jeff
  Calder}.} \bibinfo{year}{2018}\natexlab{}.
\newblock \showarticletitle{Lipschitz regularized Deep Neural Networks converge
  and generalize}.
\newblock \bibinfo{journal}{\emph{CoRR}}  \bibinfo{volume}{abs/1808.09540}
  (\bibinfo{year}{2018}).
\newblock
\showeprint[arXiv]{1808.09540}


\bibitem[Park et~al\mbox{.}(2019)]%
        {ParkFSNL19}
\bibfield{author}{\bibinfo{person}{Jeong~Joon Park}, \bibinfo{person}{Peter
  Florence}, \bibinfo{person}{Julian Straub}, \bibinfo{person}{Richard~A.
  Newcombe}, {and} \bibinfo{person}{Steven Lovegrove}.}
  \bibinfo{year}{2019}\natexlab{}.
\newblock \showarticletitle{DeepSDF: Learning Continuous Signed Distance
  Functions for Shape Representation}. In \bibinfo{booktitle}{\emph{{IEEE}
  Conference on Computer Vision and Pattern Recognition, {CVPR} 2019, Long
  Beach, CA, USA, June 16-20, 2019}}. \bibinfo{publisher}{Computer Vision
  Foundation / {IEEE}}, \bibinfo{pages}{165--174}.
\newblock


\bibitem[Poole et~al\mbox{.}(2014)]%
        {PooleSG14}
\bibfield{author}{\bibinfo{person}{Ben Poole}, \bibinfo{person}{Jascha
  Sohl{-}Dickstein}, {and} \bibinfo{person}{Surya Ganguli}.}
  \bibinfo{year}{2014}\natexlab{}.
\newblock \showarticletitle{Analyzing noise in autoencoders and deep networks}.
\newblock \bibinfo{journal}{\emph{CoRR}}  \bibinfo{volume}{abs/1406.1831}
  (\bibinfo{year}{2014}).
\newblock
\showeprint[arXiv]{1406.1831}


\bibitem[Qi et~al\mbox{.}(2017)]%
        {QiSMG17}
\bibfield{author}{\bibinfo{person}{Charles~Ruizhongtai Qi},
  \bibinfo{person}{Hao Su}, \bibinfo{person}{Kaichun Mo}, {and}
  \bibinfo{person}{Leonidas~J. Guibas}.} \bibinfo{year}{2017}\natexlab{}.
\newblock \showarticletitle{PointNet: Deep Learning on Point Sets for 3D
  Classification and Segmentation}. In \bibinfo{booktitle}{\emph{2017 {IEEE}
  Conference on Computer Vision and Pattern Recognition, {CVPR} 2017, Honolulu,
  HI, USA, July 21-26, 2017}}. \bibinfo{publisher}{{IEEE} Computer Society},
  \bibinfo{pages}{77--85}.
\newblock


\bibitem[Rakotosaona and Ovsjanikov(2020)]%
        {RakotosaonaO20}
\bibfield{author}{\bibinfo{person}{Marie{-}Julie Rakotosaona} {and}
  \bibinfo{person}{Maks Ovsjanikov}.} \bibinfo{year}{2020}\natexlab{}.
\newblock \showarticletitle{Intrinsic Point Cloud Interpolation via Dual Latent
  Space Navigation}. In \bibinfo{booktitle}{\emph{Computer Vision - {ECCV} 2020
  - 16th European Conference, Glasgow, UK, August 23-28, 2020, Proceedings,
  Part {II}}} \emph{(\bibinfo{series}{Lecture Notes in Computer Science},
  Vol.~\bibinfo{volume}{12347})}, \bibfield{editor}{\bibinfo{person}{Andrea
  Vedaldi}, \bibinfo{person}{Horst Bischof}, \bibinfo{person}{Thomas Brox},
  {and} \bibinfo{person}{Jan{-}Michael Frahm}} (Eds.).
  \bibinfo{publisher}{Springer}, \bibinfo{pages}{655--672}.
\newblock


\bibitem[Rosca et~al\mbox{.}(2020)]%
        {Mihaela2020smoothness}
\bibfield{author}{\bibinfo{person}{Mihaela Rosca}, \bibinfo{person}{Theophane
  Weber}, \bibinfo{person}{Arthur Gretton}, {and} \bibinfo{person}{Shakir
  Mohamed}.} \bibinfo{year}{2020}\natexlab{}.
\newblock \showarticletitle{A case for new neural network smoothness
  constraints}.
\newblock \bibinfo{journal}{\emph{CoRR}}  \bibinfo{volume}{abs/2012.07969}
  (\bibinfo{year}{2020}).
\newblock
\showeprint[arXiv]{2012.07969}


\bibitem[Salimans and Kingma(2016)]%
        {SalimansK16}
\bibfield{author}{\bibinfo{person}{Tim Salimans} {and}
  \bibinfo{person}{Diederik~P. Kingma}.} \bibinfo{year}{2016}\natexlab{}.
\newblock \showarticletitle{Weight Normalization: {A} Simple Reparameterization
  to Accelerate Training of Deep Neural Networks}. In
  \bibinfo{booktitle}{\emph{Advances in Neural Information Processing Systems
  29: Annual Conference on Neural Information Processing Systems 2016, December
  5-10, 2016, Barcelona, Spain}}, \bibfield{editor}{\bibinfo{person}{Daniel~D.
  Lee}, \bibinfo{person}{Masashi Sugiyama}, \bibinfo{person}{Ulrike von
  Luxburg}, \bibinfo{person}{Isabelle Guyon}, {and} \bibinfo{person}{Roman
  Garnett}} (Eds.). \bibinfo{pages}{901}.
\newblock


\bibitem[Solomon et~al\mbox{.}(2015)]%
        {SolomonGPCBNDG15}
\bibfield{author}{\bibinfo{person}{Justin Solomon}, \bibinfo{person}{Fernando
  de Goes}, \bibinfo{person}{Gabriel Peyr{\'{e}}}, \bibinfo{person}{Marco
  Cuturi}, \bibinfo{person}{Adrian Butscher}, \bibinfo{person}{Andy Nguyen},
  \bibinfo{person}{Tao Du}, {and} \bibinfo{person}{Leonidas~J. Guibas}.}
  \bibinfo{year}{2015}\natexlab{}.
\newblock \showarticletitle{Convolutional wasserstein distances: efficient
  optimal transportation on geometric domains}.
\newblock \bibinfo{journal}{\emph{{ACM} Trans. Graph.}} \bibinfo{volume}{34},
  \bibinfo{number}{4} (\bibinfo{year}{2015}), \bibinfo{pages}{66:1--66:11}.
\newblock


\bibitem[Srivastava et~al\mbox{.}(2014)]%
        {SrivastavaHKSS14}
\bibfield{author}{\bibinfo{person}{Nitish Srivastava},
  \bibinfo{person}{Geoffrey~E. Hinton}, \bibinfo{person}{Alex Krizhevsky},
  \bibinfo{person}{Ilya Sutskever}, {and} \bibinfo{person}{Ruslan
  Salakhutdinov}.} \bibinfo{year}{2014}\natexlab{}.
\newblock \showarticletitle{Dropout: a simple way to prevent neural networks
  from overfitting}.
\newblock \bibinfo{journal}{\emph{J. Mach. Learn. Res.}} \bibinfo{volume}{15},
  \bibinfo{number}{1} (\bibinfo{year}{2014}), \bibinfo{pages}{1929--1958}.
\newblock


\bibitem[Szegedy et~al\mbox{.}(2014)]%
        {SzegedyZSBEGF13}
\bibfield{author}{\bibinfo{person}{Christian Szegedy},
  \bibinfo{person}{Wojciech Zaremba}, \bibinfo{person}{Ilya Sutskever},
  \bibinfo{person}{Joan Bruna}, \bibinfo{person}{Dumitru Erhan},
  \bibinfo{person}{Ian~J. Goodfellow}, {and} \bibinfo{person}{Rob Fergus}.}
  \bibinfo{year}{2014}\natexlab{}.
\newblock \showarticletitle{Intriguing properties of neural networks}. In
  \bibinfo{booktitle}{\emph{2nd International Conference on Learning
  Representations, {ICLR} 2014, Banff, AB, Canada, April 14-16, 2014,
  Conference Track Proceedings}}, \bibfield{editor}{\bibinfo{person}{Yoshua
  Bengio} {and} \bibinfo{person}{Yann LeCun}} (Eds.).
\newblock


\bibitem[Terj{\'{e}}k(2020)]%
        {Terjek20}
\bibfield{author}{\bibinfo{person}{D{\'{a}}vid Terj{\'{e}}k}.}
  \bibinfo{year}{2020}\natexlab{}.
\newblock \showarticletitle{Adversarial Lipschitz Regularization}. In
  \bibinfo{booktitle}{\emph{8th International Conference on Learning
  Representations, {ICLR} 2020, Addis Ababa, Ethiopia, April 26-30, 2020}}.
  \bibinfo{publisher}{OpenReview.net}.
\newblock


\bibitem[Tibshirani(1996)]%
        {tibshirani1996regression}
\bibfield{author}{\bibinfo{person}{Robert Tibshirani}.}
  \bibinfo{year}{1996}\natexlab{}.
\newblock \showarticletitle{Regression shrinkage and selection via the lasso}.
\newblock \bibinfo{journal}{\emph{Journal of the Royal Statistical Society:
  Series B (Methodological)}} \bibinfo{volume}{58}, \bibinfo{number}{1}
  (\bibinfo{year}{1996}), \bibinfo{pages}{267--288}.
\newblock


\bibitem[Tihonov(1963)]%
        {tihonov1963solution}
\bibfield{author}{\bibinfo{person}{Andrei~Nikolajevits Tihonov}.}
  \bibinfo{year}{1963}\natexlab{}.
\newblock \showarticletitle{Solution of incorrectly formulated problems and the
  regularization method}.
\newblock \bibinfo{journal}{\emph{Soviet Math.}}  \bibinfo{volume}{4}
  (\bibinfo{year}{1963}), \bibinfo{pages}{1035--1038}.
\newblock


\bibitem[Ulyanov et~al\mbox{.}(2018)]%
        {ulyanov2018deep}
\bibfield{author}{\bibinfo{person}{Dmitry Ulyanov}, \bibinfo{person}{Andrea
  Vedaldi}, {and} \bibinfo{person}{Victor Lempitsky}.}
  \bibinfo{year}{2018}\natexlab{}.
\newblock \showarticletitle{Deep image prior}. In
  \bibinfo{booktitle}{\emph{Proceedings of the IEEE conference on computer
  vision and pattern recognition}}. \bibinfo{pages}{9446--9454}.
\newblock


\bibitem[Varga et~al\mbox{.}(2017)]%
        {varga2017gradient}
\bibfield{author}{\bibinfo{person}{D{\'a}niel Varga},
  \bibinfo{person}{Adri{\'a}n Csisz{\'a}rik}, {and} \bibinfo{person}{Zsolt
  Zombori}.} \bibinfo{year}{2017}\natexlab{}.
\newblock \showarticletitle{Gradient regularization improves accuracy of
  discriminative models}.
\newblock \bibinfo{journal}{\emph{arXiv preprint arXiv:1712.09936}}
  (\bibinfo{year}{2017}).
\newblock


\bibitem[Virmaux and Scaman(2018)]%
        {VirmauxS18}
\bibfield{author}{\bibinfo{person}{Aladin Virmaux} {and} \bibinfo{person}{Kevin
  Scaman}.} \bibinfo{year}{2018}\natexlab{}.
\newblock \showarticletitle{Lipschitz regularity of deep neural networks:
  analysis and efficient estimation}. In \bibinfo{booktitle}{\emph{Advances in
  Neural Information Processing Systems 31: Annual Conference on Neural
  Information Processing Systems 2018, NeurIPS 2018, December 3-8, 2018,
  Montr{\'{e}}al, Canada}}, \bibfield{editor}{\bibinfo{person}{Samy Bengio},
  \bibinfo{person}{Hanna~M. Wallach}, \bibinfo{person}{Hugo Larochelle},
  \bibinfo{person}{Kristen Grauman}, \bibinfo{person}{Nicol{\`{o}}
  Cesa{-}Bianchi}, {and} \bibinfo{person}{Roman Garnett}} (Eds.).
  \bibinfo{pages}{3839--3848}.
\newblock


\bibitem[Wang et~al\mbox{.}(2018)]%
        {WangZLFLJ18}
\bibfield{author}{\bibinfo{person}{Nanyang Wang}, \bibinfo{person}{Yinda
  Zhang}, \bibinfo{person}{Zhuwen Li}, \bibinfo{person}{Yanwei Fu},
  \bibinfo{person}{Wei Liu}, {and} \bibinfo{person}{Yu{-}Gang Jiang}.}
  \bibinfo{year}{2018}\natexlab{}.
\newblock \showarticletitle{Pixel2Mesh: Generating 3D Mesh Models from Single
  {RGB} Images}. In \bibinfo{booktitle}{\emph{Computer Vision - {ECCV} 2018 -
  15th European Conference, Munich, Germany, September 8-14, 2018, Proceedings,
  Part {XI}}} \emph{(\bibinfo{series}{Lecture Notes in Computer Science},
  Vol.~\bibinfo{volume}{11215})}, \bibfield{editor}{\bibinfo{person}{Vittorio
  Ferrari}, \bibinfo{person}{Martial Hebert}, \bibinfo{person}{Cristian
  Sminchisescu}, {and} \bibinfo{person}{Yair Weiss}} (Eds.).
  \bibinfo{publisher}{Springer}, \bibinfo{pages}{55--71}.
\newblock


\bibitem[Weng et~al\mbox{.}(2018)]%
        {WengZCSHDBD18}
\bibfield{author}{\bibinfo{person}{Tsui{-}Wei Weng}, \bibinfo{person}{Huan
  Zhang}, \bibinfo{person}{Hongge Chen}, \bibinfo{person}{Zhao Song},
  \bibinfo{person}{Cho{-}Jui Hsieh}, \bibinfo{person}{Luca Daniel},
  \bibinfo{person}{Duane~S. Boning}, {and} \bibinfo{person}{Inderjit~S.
  Dhillon}.} \bibinfo{year}{2018}\natexlab{}.
\newblock \showarticletitle{Towards Fast Computation of Certified Robustness
  for ReLU Networks}. In \bibinfo{booktitle}{\emph{Proceedings of the 35th
  International Conference on Machine Learning, {ICML} 2018,
  Stockholmsm{\"{a}}ssan, Stockholm, Sweden, July 10-15, 2018}}
  \emph{(\bibinfo{series}{Proceedings of Machine Learning Research},
  Vol.~\bibinfo{volume}{80})}, \bibfield{editor}{\bibinfo{person}{Jennifer~G.
  Dy} {and} \bibinfo{person}{Andreas Krause}} (Eds.).
  \bibinfo{publisher}{{PMLR}}, \bibinfo{pages}{5273--5282}.
\newblock


\bibitem[Williams et~al\mbox{.}(2019)]%
        {WilliamsSSZBP19}
\bibfield{author}{\bibinfo{person}{Francis Williams}, \bibinfo{person}{Teseo
  Schneider}, \bibinfo{person}{Cl{\'{a}}udio~T. Silva}, \bibinfo{person}{Denis
  Zorin}, \bibinfo{person}{Joan Bruna}, {and} \bibinfo{person}{Daniele
  Panozzo}.} \bibinfo{year}{2019}\natexlab{}.
\newblock \showarticletitle{Deep Geometric Prior for Surface Reconstruction}.
  In \bibinfo{booktitle}{\emph{{IEEE} Conference on Computer Vision and Pattern
  Recognition, {CVPR} 2019, Long Beach, CA, USA, June 16-20, 2019}}.
  \bibinfo{publisher}{Computer Vision Foundation / {IEEE}},
  \bibinfo{pages}{10130--10139}.
\newblock


\bibitem[Williams et~al\mbox{.}(2021)]%
        {WilliamsTBZ21}
\bibfield{author}{\bibinfo{person}{Francis Williams}, \bibinfo{person}{Matthew
  Trager}, \bibinfo{person}{Joan Bruna}, {and} \bibinfo{person}{Denis Zorin}.}
  \bibinfo{year}{2021}\natexlab{}.
\newblock \showarticletitle{Neural Splines: Fitting 3D Surfaces With
  Infinitely-Wide Neural Networks}. In \bibinfo{booktitle}{\emph{{IEEE}
  Conference on Computer Vision and Pattern Recognition, {CVPR} 2021, virtual,
  June 19-25, 2021}}. \bibinfo{publisher}{Computer Vision Foundation / {IEEE}},
  \bibinfo{pages}{9949--9958}.
\newblock


\bibitem[Xie et~al\mbox{.}(2021)]%
        {neuralfieldsurvey}
\bibfield{author}{\bibinfo{person}{Yiheng Xie}, \bibinfo{person}{Towaki
  Takikawa}, \bibinfo{person}{Shunsuke Saito}, \bibinfo{person}{Or Litany},
  \bibinfo{person}{Shiqin Yan}, \bibinfo{person}{Numair Khan},
  \bibinfo{person}{Federico Tombari}, \bibinfo{person}{James Tompkin},
  \bibinfo{person}{Vincent Sitzmann}, {and} \bibinfo{person}{Srinath Sridhar}.}
  \bibinfo{year}{2021}\natexlab{}.
\newblock \showarticletitle{Neural Fields in Visual Computing and Beyond}.
\newblock \bibinfo{journal}{\emph{CoRR}}  \bibinfo{volume}{abs/2111.11426}
  (\bibinfo{year}{2021}).
\newblock


\bibitem[Yoshida and Miyato(2017)]%
        {YoshidaM17}
\bibfield{author}{\bibinfo{person}{Yuichi Yoshida} {and}
  \bibinfo{person}{Takeru Miyato}.} \bibinfo{year}{2017}\natexlab{}.
\newblock \showarticletitle{Spectral Norm Regularization for Improving the
  Generalizability of Deep Learning}.
\newblock \bibinfo{journal}{\emph{CoRR}}  \bibinfo{volume}{abs/1705.10941}
  (\bibinfo{year}{2017}).
\newblock
\showeprint[arXiv]{1705.10941}


\end{thebibliography}
\clearpage{}
\appendix

\section{Experimental \& Implementation Details}
For all our experiments, we initialize the per-layer Lipschitz constant $c_i = \| \mW_i \|_\infty$ in \refequ{add_regularization} as the Lipschitz constant of the initial weight matrix. The weight matrices are initialized with the method by \cite{GlorotB10} if the activation is $\tanh$ and with the method by \citet{HeZRS15} if the activation is ReLU or its variants.
In the following subsections, we present details of individual experiments in the main text.

\subsection{3D Neural Implicit Interpolation}\label{app:3D_interpolation}
We use the DeepSDF architecture \cite{ParkFSNL19} to design the interpolation experiment of 3D neural SDFs (\reffig{teaser}, \reffignum{3d_interpolation}, \reffignum{multiSDF_interp}). The inputs to our network are the query location in $\R^3$ and the latent code $t$. Our network consists of 5 hidden layers of 256 neurons with the $\tanh$ activation. We multiply the input point $x$ with one hundred $100x$ to avoid the possibility that the network smoothness in the latent code is bounded by spatial smoothness. The last layer is a linear layer outputting the signed distance value at the input query location. 
The training shapes are normalized to the bounding box between 0 and 1. We use the mean square error (MSE) as our loss function $\mathcal{L}(\theta)$ for our baseline model. We augment the MSE loss with our Lipschitz regularization \refequ{add_regularization} (with $α = 10^{-6}$) to evaluate the influence of our method. We compute the MSE loss by sampling $10^5$ points where $40\%$ of them are on the surface, $40\%$ are near the surface, and $20\%$ are drawn from uniformly sampling the bounding box.
We use the Adam optimizer \cite{KingmaB14} with its default parameters presented and learning rate $10^{-4}$.

For our experiment on occupancy interpolation \cite{MeschederONNG19} (\reffig{occupancy_interp}), we use the same architecture as our SDF experiments and append a \emph{sigmoid} function right before the output. Our loss function is the cross-entropy loss with and without our Lipschitz regularization (with $α = 10^{-6}$).

In these interpolation experiments, we pre-determine the latent codes for each shape we want to interpolate. For example \reffig{teaser}, we minimize the loss with respect to the SDF of a torus when $t=0$ and with the double torus when $t=1$. Similarly, we also use $t = 0, 1$ in our occupancy interpolation \reffig{occupancy_interp}. In \reffig{3d_interpolation}, the three sets of code for the green shapes are [0,0], [1,0], [0.5, 0.866]. In \reffig{multiSDF_interp}, the latent code for the four shapes are set to be one-hot vectors, such as [1,0,0,0]. 

\subsection{2D Neural Implicit Interpolation}\label{app:2D_interpolation}
The experiments on interpolating 2D neural implicits are similar to the above-mentioned 3D interpolation ones \refapp{3D_interpolation}. We minimize the MSE loss with Adam and use the DeepSDF architecture. The only difference is that the network is smaller and our training data is sampled uniformly on the 2D space.
In \reffig{dirichlet}, \reffignum{manual_lipc_large}, we use a MLP with 5 hidden layers of 64 neurons with ReLU activation. Specifically, we set the Dirichlet regularization \refequ{dirichlet_regularization} to be $10^{-4}$ in \reffig{dirichlet}. We manually set the Lipschitz constant per layer to be $1.4$ in \reffig{manual_lipc_large}. For our Lipschitz regularization, we set $α = 3 × 10^{-6}$.

\subsection{Toy 2D test time optimization}
In \reffig{toy_testtime_adam}, we present a toy test time optimization to demonstrate the importance of having a smooth latent space. We use the same training set-up as \refapp{2D_interpolation} to train our interpolation networks. During test-time, we initialize the latent code to be $t = 0.5$ and we randomly sample 8 points on the iso-line of the star shape and minimize the square distance at these sample points. Intuitively, if the loss is minimized, these points will lie on the zero iso-line of the optimized SDF. In \reffig{toy_testtime_adam}, we show the optimization using Adam. We also tried to optimize the code using SGD, but we only notice a small difference in this toy set-up.
\begin{figure}
  \begin{center}
  \includegraphics[width=1\linewidth]{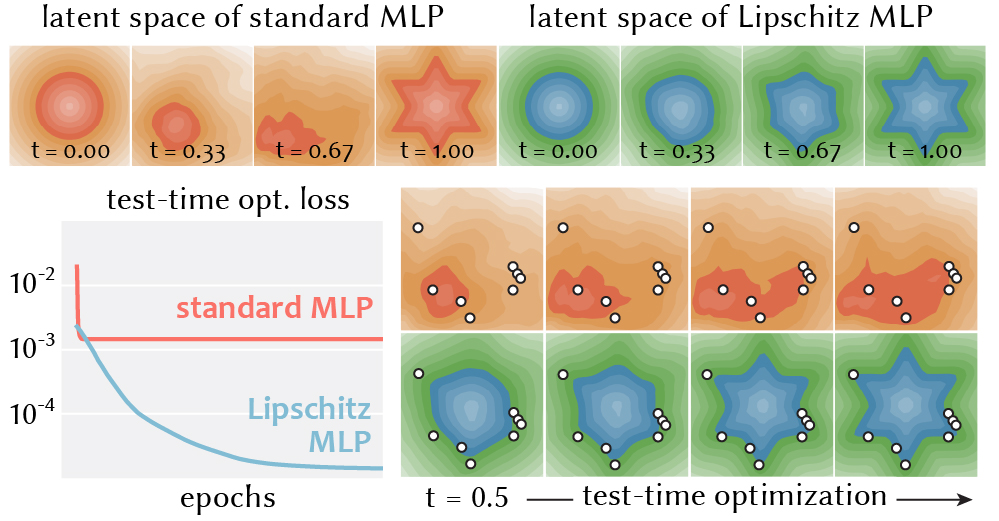}
  \end{center}
  \vspace{-5pt}
  \caption{In this simple toy example, test-time optimization using SGD gives us similar result compared to the one optimized with Adam (see \reffig{toy_testtime_adam}).}
  \label{fig:toy_testtime}
\end{figure}
\begin{figure*}
  \includegraphics[width=\linewidth, trim={0mm 5mm 0mm 0mm}]{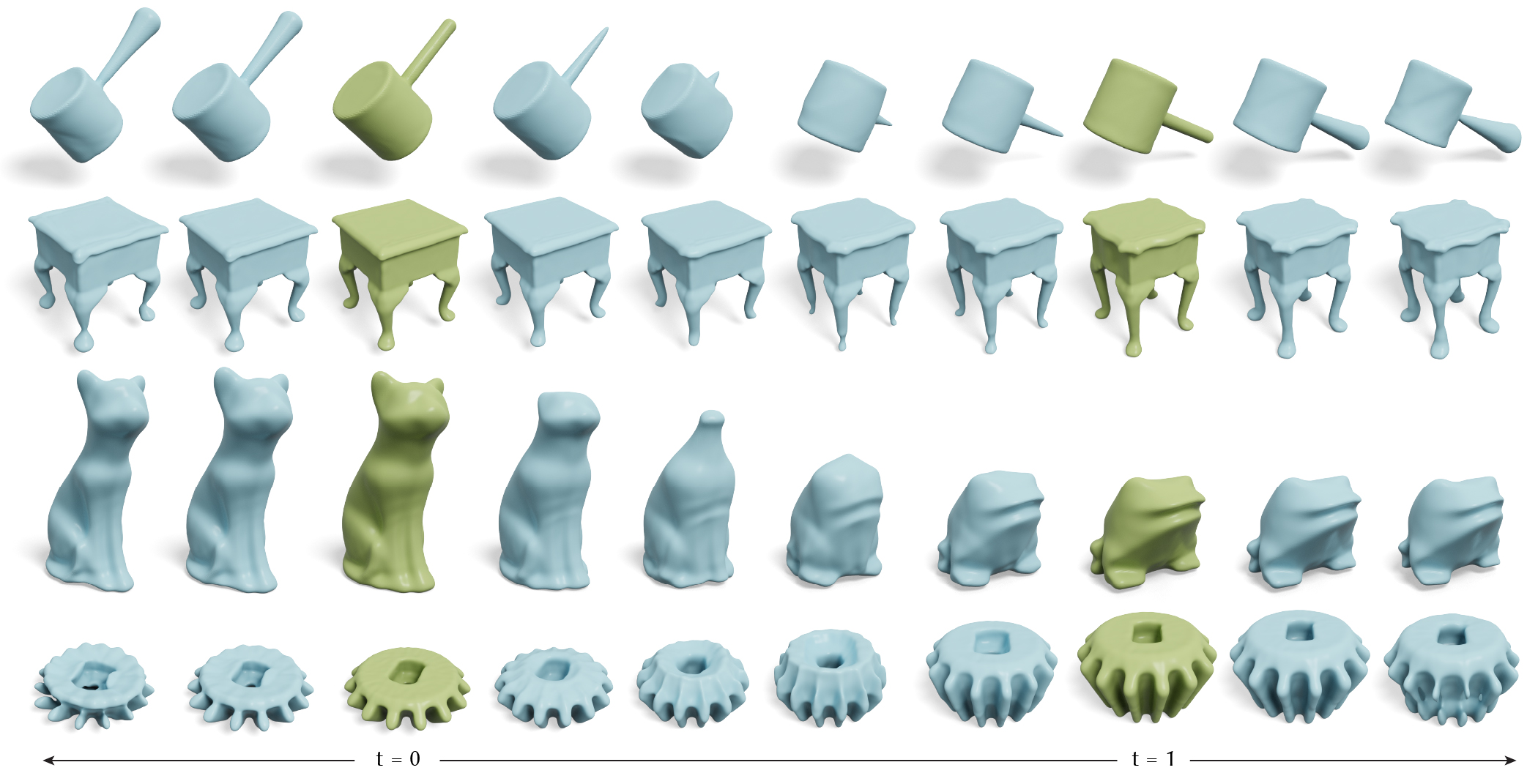}
  \vspace*{-0.5\intextsep}
  \caption{\update{We fit neural networks to the signed distance field of a shape when the latent code $t = 0$ and another shape when $t = 1$ (green). Our Lipschitz regularization encourages smooth interpolation and extrapolation (blue) even when trained on only a pair of shapes.}}
  \label{fig:more_sdf_interp} 
\end{figure*}

\subsection{Test time optimization}\label{app:testtime_optimization}
We evaluate test time optimization on the chair category of the ShapeNet dataset \cite{shapenet}, which contains 4746 shapes.
\update{Our network architecture follows the baseline autoencoder model in \cite{duggal2022mending}. The encoder is a PointNet \cite{QiSMG17}. Specifically, the input to our PointNet is a point location $\vp_i$ in $\R^3$. We pass this point $\vp_i$ to a fully-connected network of size [3, 256, 512] with $\tanh$ activation to transform this point to a feature vector $\vz_i$ of size 512. This fully-connected network is used to independently process each point $\vp_i$ in the point cloud. Thus, after this step, we obtain a $n$-by-512 feature matrix where $n$ is the number of points. We perform a max-pooling for this feature matrix to obtain a feature vector $\vz_\text{global}$ of size 512, encoding the global information of the point cloud. We then concatenate this global feature vector $\vz_\text{global}$ is with the feature vector of individual point feature $\vz_i$. After concatenation, we pass this local-global feature [$\vz_i$, $\vz_\text{global}$] to the second fully-connected network of size [1024, 512, 256] with $\tanh$ activation. Similar to the previous step, we use this shared network to process each point independently. Thus, after processing all the points, we will obtain a $n$-by-256 feature matrix. We then perform another max-pooling to turn this $n$-by-256 feature matrix into a 256 global feature vector. To prevent the latent codes from diverging to an arbitrarily vector with large magnitude, apply a sigmoid function to the global feature vector ensure each latent code lies between 0 and 1. We then treat this as the final feature representation of the point cloud after the encoding process.} 

\update{After the encoding process, our decoder is a DeepSDF \cite{ParkFSNL19} which takes the query location in $\R^3$ and the 256 latent vector from the encoder as its input, and then outputs the signed distance value.} The decoder has size [259, 1024, 1024, 1024, 512, 256, 128, 1] with the leaky ReLU activation. 
%
%
Similar to \refapp{3D_interpolation}, we minimize the MSE loss with Adam and set the learning rate to be $10^{-4}$. To evaluate our method, we add a Lipschitz regularization with $α = 10^{-11}$ to the decoder during training (see \reffig{testtime_qualitative}).

For the test time optimization, given a partial point cloud, we initialize the latent code by passing through our PointNet encoder. With this code, we pass each point in the point cloud to the decoder and minimize the square SDF value. Intuitively, we want the point on the partial point cloud to lie on the zero iso-surface. In addition, we also augment an Eikonal term \cite{GroppYHAL20} weighted by $1e-2$ to encourage the output to be an SDF-like function. 
We minimize this loss (square SDF and an Eikonal loss) by changing the latent code parameter (the parameter before applying the sigmoid) during test time with Adam with a learning rate $10^{-4}$ until converged. 

\subsection{Alternative Lipschitz Regularizations}
We evaluate our method against the method proposed in \cite{AnilLG19} (\refequ{james}) and another alternative mentioned in \refequ{direct_weight} on 2D interpolation tasks \refapp{2D_interpolation}. We use a 5 layer ReLU MLP with 64 neurons on each hidden layer. Because these approaches are all defined on the Lipschitz bound of the network, we use the same $α = 10^{-6}$ for a fair comparison. 

We also compare against the method by \citet{YoshidaM17} on 2D interpolation. We use 5 and 10 layers ReLU MLP with 64 neurons on each hidden layer respectively. We use $α = 10^{-5}$ for the method by \citet{YoshidaM17} and $α = 10^{-6}$ for our regularization.

In \refequ{log_lipschitz}, we evaluate it on a large network trained on the ShapeNet \cite{shapenet}. We notice that if the task is simple, such as 2D interpolation, whether to take a log result in similar performance when we have a good $α$. But when evaluating on large experiments with large networks, minimizing the log of the Lipschitz bound \refequ{log_lipschitz} may start to have issues on convergence. In this experiment specifically, we use the same training set-up as \refapp{testtime_optimization} and we use $α = 10^{-6}$.

\begin{figure*}
  \includegraphics[width=\linewidth, trim={0mm 5mm 0mm 0mm}]{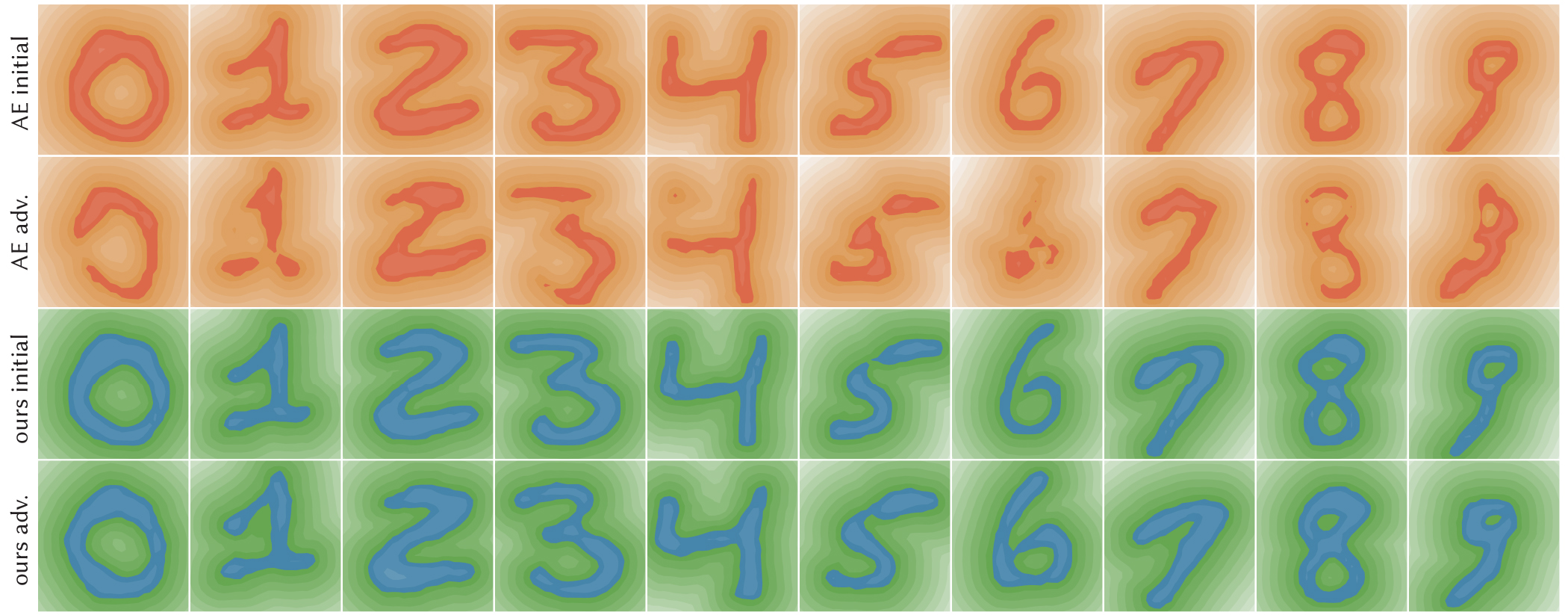}
  \vspace*{-0.5\intextsep}
  \caption{\update{We perform adversarial attacks in the latent space, the same setup as \reffig{virtual_adversarial}. We can observe that Vanilla AE is vulnerable to the attack so the initial SDFs (first row) are completely destroyed after adversarial perturbations (second row). In contrast, our Lipschitz regularized network is more robust to the attack (third and the fourth rows). }}
  \label{fig:more_adversaries} 
\end{figure*}
\subsection{MNIST Implicit Autoencoder}\label{app:MNIST}
The experiments presented in \refsec{weight_decay} and \refsec{robustness} are evaluated on the MNIST dataset (60000 hand-written digits) represented in 28-by-28 SDF images.
Our autoencoder uses two MLPs as our encoder and decoder. Our encoder has size [784, 256, 128, 64, 32] with leaky ReLU activation. It takes the image of an MNIST digit (in SDF form) as the input and outputs a latent code with dimension 32. Similar to \refapp{testtime_optimization}, we then apply a sigmoid function on the output to ensure the actual latent code lies between 0 and 1. 
The inputs to our decoder are the latent code and the position in the image space. It outputs the SDF value at the location. Our decoder has dimension [35, 128, 128, 128, 1] with the sorting activation \cite{AnilLG19}. We multiply the input position by 100 to avoid the possibility that the network is constrained by spatial smoothness
We use Adam with a learning rate $10^{-4}$ to minimize the MSE loss evaluated on the 28-by-28 regular 2D grid. 
For each regularization (L1, L2, and ours), we perform parameter sweeping on log scale and report the best one in terms of test accuracy. Specifically, we use $10^{-7}$ for both the L1 and L2 regularization, and $10^{-6}$ for our Lipschitz regularization.

\update{To construct an adversarial perturbation in the latent space, we fist obtain the initial latent code $\vt_i$ of a valid MNIST digit by passing an image $\mI_i$ from the training/testing set to our encoder, then we follow the \emph{fast gradient signed method} proposed in \cite{GoodfellowSS14} to compute the adversarial perturbation. Specifically, we set the loss function $\mathcal{J}$ to be squared L2 pixel difference between the input MNIST digit $\mI_i$ and the decoded image output by the network $f_θ(\vt)$ using another code $\vt$. Then the adversarial latent code is constructed by
\begin{align}
  \vt_\text{adv.} = \vt_i + ϵ\ \text{sign}\Big(\frac{∂ \mathcal{J}(\mI_i, f_θ(\vt))}{∂ \vt}\Big)
\end{align}
with predetermined small magnitude $ϵ = 0.05$. Then the adversarial MNIST digits can be obtained by visualizing the output of the network with the adversarial code $f_\theta(\vt_\text{adv.})$.}

\begin{figure}
  \begin{center}
  \includegraphics[width=1\linewidth]{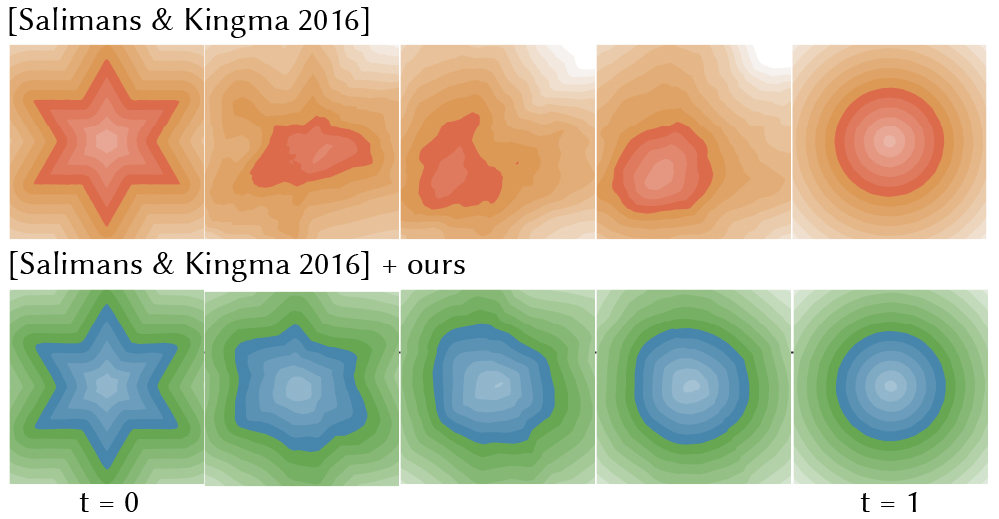}
  \end{center}
  \vspace{-5pt}
  \caption{\update{Our Lipschitz regularization can complement other reparameterization schemes, such as the \emph{weight normalization} by \citet{SalimansK16}. We show that solely with weight normalization is insufficent for obtain smooth results (top row), but our method can be used jointly with \cite{SalimansK16} to obtain smooth interpolation (bottom row).}}
  \label{fig:weight_normalization_compare}
\end{figure}

\update{
\section{Relationship with Weight Normalization}
\emph{Weight Normalization} is a reparameterization technique proposed by \citet{SalimansK16} to accelerate the training process. The key idea is to parameterize the weight matrix $\mW$ with a trainable matrix $\mV$ and a trainable scaling factor $g$ 
\begin{align}
  \mW = g × \frac{\mV}{\| \mV\|}
\end{align} 
where the matrix $\mV$ is normalized to have unit norm and $g$ is the scaling factor that controls the magnitude of $\mW$. This reparameterization is similar to our weight normalization layer in \refsec{normalization_layer}, but with a different norm. However, the key difference is that this reparameterization along is insufficient to guarantee smoothness. In the first row of \reffig{weight_normalization_compare}, we can observe that solely with the method by \citet{SalimansK16} still results in non-smooth interpolation. This is because there is no regularization to encourage small $g$. In the second row of \reffig{weight_normalization_compare}, we demonstrate the flexibility of our method that we can apply our Lipschitz regularization to encourage small $g$ under the reparameterization in \cite{SalimansK16} and successfully lead to smooth interpolation results.
}

\begin{figure}
  \begin{center}
  \includegraphics[width=1\linewidth]{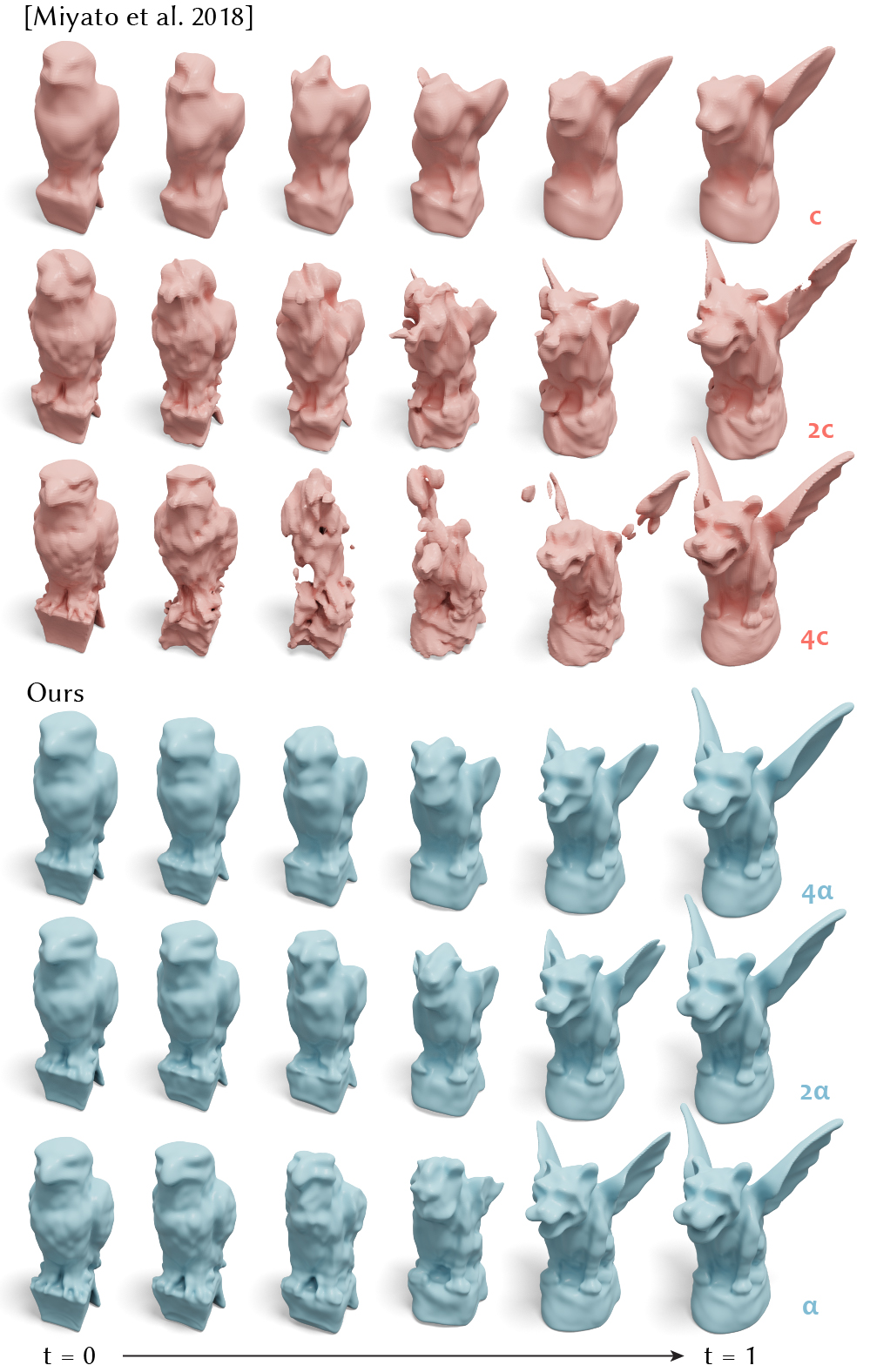}
  \end{center}
  \vspace{-5pt} 
  \caption{\update{Lipschitz constrained networks, such as \cite{MiyatoKKY18}, are sensitive to the change of the prescribed Lipschitz bound. We can observe a dramatic change from too smooth (1st row) to too non-smooth (3rd row) with a minor logarithmic scaling of the initial Lipschitz bound $c$. In contrast, our method (blue) is more robust with respect to the change of our regularization weight $\alpha$.}}
  \label{fig:3D_SN}
\end{figure}

\update{
\section{Comparison with Spectral Normalization}
\emph{Spectral Normalization} proposed by \citet{MiyatoKKY18} is a method to constrain the Lipschitz bound of a network. As discussed in \refsec{relatedwork}, these Lipschitz constrained networks are sensitive to the choice of the Lipschitz bound. In most geometry applications, a good choice of bound is unknown, thus it requires extensive hyperparameter tuning. In \reffig{3D_SN}, we show that the results of Lipschitz constrained networks change dramatically when increasing the bounds. Our method, instead, results in smoother change with better results when playing with our regularization parameter $\alpha$.}

\end{document}